\documentclass[10pt,journal,compsoc]{IEEEtran}
\usepackage[dvipsnames]{xcolor}

\usepackage{multirow}
\usepackage{bigstrut}

\ifCLASSOPTIONcompsoc
  \usepackage[nocompress]{cite}
\else
  \usepackage{cite}
\fi
\ifCLASSINFOpdf
\usepackage[pdftex]{graphicx}
\else
\fi
\usepackage{amsmath}
\usepackage{algorithmic}

\usepackage{array}

\usepackage{verbatim}
\usepackage[utf8]{inputenc}
\usepackage{soul}
\usepackage{array,booktabs}

\usepackage{pifont}
\newcommand{\cmark}{\color{black}\ding{52}}%
\newcommand{\xmark}{\color{black}\ding{56}}%

\newcommand{\figref}[1]{Fig.~\ref{#1}}%
\newcolumntype{P}[1]{>{\centering\arraybackslash}p{#1}}
\newcolumntype{M}[1]{>{\centering\arraybackslash}m{#1}}

\begin{document}
%
\title{A Survey of Graph-based Deep Learning for Anomaly Detection in Distributed Systems}
%
%
%
%

\author{{Armin Danesh Pazho* \thanks{* First two authors (A. Danesh Pazho and G. Alinezhad Noghre) have equal contribution.}, Ghazal Alinezhad Noghre*, Arnab A Purkayastha, Jagannadh Vempati, Otto Martin, and Hamed Tabkhi}
\IEEEcompsocitemizethanks{\IEEEcompsocthanksitem Armin Danesh Pazho, Ghazal Alinezhad Noghre, and Hamed Tabkhi are with the University of North Carolina at Charlotte,
NC, USA \protect\\
E-mail: adaneshp@uncc.edu, galinezh@uncc.edu, and htabkhiv@uncc.edu respectively.
\IEEEcompsocthanksitem Arnab A Purkayastha is with Western New England University, MA USA \protect\\
E-mail: arnab.purkayastha@wne.edu.
\IEEEcompsocthanksitem Jagannadh Vempati and Martin Otto are with Siemens Technology, Princeton, NJ, USA. \protect\\
E-mail: jagannadh.vempati@siemens.com and m.otto@siemens.com respectively}
}

%
%

\markboth{A. Danesh Pazho et al.}%
{Shell \MakeLowercase{\textit{et al.}}: A Survey of Graph-based Deep Learning for Anomaly Detection in Distributed Systems}
%



\IEEEtitleabstractindextext{%

\begin{abstract}

Anomaly detection is a crucial task in complex distributed systems. A thorough understanding of the requirements and challenges of anomaly detection is pivotal to the security of such systems, especially for real-world deployment. While there are many works and application domains that deal with this problem, few have attempted to provide an in-depth look at such systems. In this survey, we explore the potentials of graph-based algorithms to identify anomalies in distributed systems. These systems can be heterogeneous or homogeneous, which can result in distinct requirements. One of our objectives is to provide an in-depth look at graph-based approaches to conceptually analyze their capability to handle real-world challenges such as heterogeneity and dynamic structure. This study gives an overview of the State-of-the-Art (SotA) research articles in the field and compare and contrast their characteristics. To facilitate a more comprehensive understanding, we present three systems with varying abstractions as use cases. We examine the specific challenges involved in anomaly detection within such systems. Subsequently, we elucidate the efficacy of graphs in such systems and explicate their advantages. We then delve into the SotA methods and highlight their strength and weaknesses, pointing out the areas for possible improvements and future works.

\end{abstract}

}

\maketitle

\IEEEdisplaynontitleabstractindextext

%
\IEEEpeerreviewmaketitle

\IEEEraisesectionheading{\section{Introduction}\label{sec:introduction}}
\IEEEPARstart{A}{nomaly} detection refers to finding abnormal behavior or patterns in the data or a system that does not match the expected behavior \cite{chandola2009anomaly}. In other words, a non-benign change in the known and correct behavior of a system can be detected as an anomaly. Anomaly detection is critical in both homogeneous and heterogeneous distributed systems. In heterogeneous distributed systems, different components, from small variations like sensors, work all the way up to large components like control facilities to achieve the overall goal of the system. All these components are spread out in a big network; hence the word distributed, and they are dissimilar with respect to their structure and data production, hence the word heterogeneous. Homogeneous systems can be seen as a special case of heterogeneous systems where all the components are very similar if not identical.

Many distributed systems are operational in critical and/or important fields (e.g., Power Generation and Distribution systems). Thus, their security and correctness are critical. Vulnerabilities can be seen in every aspect of these systems and in different levels of abstraction; from problematic individual components within a system or a network to seeing abnormalities or irregularities where these components connect with each other to accomplish a more advanced task. Table \ref{ApplicationsTable} shows several distributed networks and possible sources of anomalies in them. Anomaly detection in these systems must meet many requirements, making the task even more complicated. Anomalies must be detected by processing large quantities of data, considering the nature of anomalies, which makes detecting them a challenging task and the real-world setup limitations.

\begin{table}[ht]
\caption{Example of distributed systems with sample source of anomalies}
\renewcommand{\arraystretch}{1.15}
\centering
\begin{tabular}{>{\centering\arraybackslash}m{0.4\linewidth}>{\centering\arraybackslash}m{0.5\linewidth}}
\textbf{Network}                  & \textbf{Sources of Anomaly}                                                 \\ \hline \hline
Power Generation and Distribution & Anomalies in hardware,  software, and the network                           \\ \hline
Smart Video Surveillance          & Abnormal and dangerous events, hardware or software issues                  \\ \hline
Smart Cities                      & Fault in Infrastructure, safety and privacy issues                          \\ \hline
Smart Grids                       & Faults, cyber-attacks, natural disturbances                                 \\ \hline
Social Network                    & Illegal activities, bullying, spams, offensive content                      \\ \hline
Telecommunication                 & Malfunctions, hardware issues, intrusion                                    \\ \hline
Company’s Internal Network        & Hardware and software issues, unwanted access, intrusion                    \\ \hline
Factories’ Production Line        & Fault in manufacturing, problem in manufacturing line equipment             \\ \hline
Smart Transportation              & Technical issues, hardware and software issues, abnormal trajectory changes \\ \hline
Banking System                    & Money laundering, fraud transactions, intrusion                     
\end{tabular}
\label{ApplicationsTable}
\vspace{-10pt}
\end{table}

In recent years, neural networks have shown great potential and are being explored extensively. They have opened new avenues and increased interest in extending deep learning approaches for anomaly detection \cite{pang2021deep, 10.1145/3437963.3441659}. Traditional methods are mostly dependent on handcrafted features defined by domain experts which makes them less generalizable and flexible for different domains. Feature selection can be time-consuming and error-prone. In contrast, deep learning methods are highly adaptable to the specific requirements of each domain. Deep learning models help with automation as they need minimal effort and supervision for identifying the prominent features without manual feature selection. They are more suitable for learning complex information, their result is more generalizable and less prone to over-fitting, and they are model free. Another advantage of deep learning algorithms in the context of anomaly detection is that they can handle large input data more efficiently and scale up compared to traditional methods. This characteristic makes them ideal for anomaly detection in complex distributed systems. Deep learning-based algorithms have proved to be more effective and generally show better results. This is due to the fact that they are capable of building more enriched features and detecting complex patterns compared to traditional methods.

Real-world distributed systems are an intertwined network of numerous components, each of which representing distinctive properties, changing over time. Graphs prove to be advantageous in capturing the relational dynamic of these components as well as their individual features. Mapping a system to nodes and edges of a graph, allows for a better comprehension of the system. It is noteworthy that not all types of graphs are capable of capturing every aspect of a system. While conventional graph representations are adequate for homogeneous systems, heterogeneous systems require the more complex attributed graphs. Attributed graph is a type of data structure where each node and edge is associated with a set of attributes or properties. In heterogeneous systems, they allow for a the representation of multiple variations of entities and relationships, each with their own unique attributes \cite{wang2021survey}. On the other hand, to address the changing nature of distributed systems, dynamic graphs come to aid. In dynamic graphs, nodes and edges can appear, disappear, or change over time, rendering them ideal for dynamic systems. Hence, When surveying graph-based methods, we analyze their ability to handle attributed and dynamic graphs.

Numerous studies have reviewed techniques and tools for general anomaly detection problems \cite{chalapathy2019deep, pang2021deep, ruff2021unifying, fernandes2019comprehensive, 10.1145/3437963.3441659}. Some of the existing surveys focus on anomaly detection in big data such as \cite{habeeb2019real, thudumu2020comprehensive}. In addition, certain studies investigate methods utilized for anomaly detection in specific narrow domains such as fake news detection \cite{ahmed2022combining}, social media interactions \cite{10.1145/2980765.2980767}, financial transactions \cite{ahmed2016survey}, etc. Several articles concentrate on the applications of graphs for identifying anomalies in different systems. \cite{ranshous2015anomaly} overviews anomaly detection algorithms in graphs of dynamic networks; however it mostly mentions the traditional methods and neglects state-of-the-art deep learning methods without mentioning the challenges and difficulties of anomaly detection in such systems. A recent study \cite{ma2021comprehensive} tries to fill this gap and focuses on deep learning methods for anomaly detection in graphs. \cite{ma2021comprehensive} also mentions the challenges in this area, but falls short in categorizing the challenges and analyzing the real-world requirements and constraints. Both \cite{ranshous2015anomaly, ma2021comprehensive} do not provide different use cases and analyze them with respect to review algorithms. Another work \cite{akoglu2015graph} provides a general overview of methods for anomaly detection in data represented as graphs and discusses the challenges and use cases of these methods. However, this work did not focus on deep learning methods for anomaly detection, as these techniques were not yet widely established at the time of the study.

The major focus of this survey is to provide a comprehensive overview of the state-of-art graph-based techniques to solve the problem of anomaly detection. In particular, we look at real-time complex distributed systems and qualitatively model those to identify and analyze various methods in anomaly detection that utilize the benefits of graphs. In a nutshell, we make the following notable contributions to this survey:

\begin{itemize}
    \item Modeling three conceptual use cases and employing them for exploring the requirements, challenges, and benefits of anomaly detection algorithms.
    \item Identifying the logical, algorithmic, and implementation requirements and challenges of anomaly detection in real-world distributed systems.
    \item Comparing and contrasting the different characteristics of the graph-based anomaly detection algorithms, tools, and techniques in-depth for dealing with anomalies and qualitatively modeling them over the three conceptual use cases that we introduce.
    \item Providing future research direction for graph-based anomaly detection in distributed systems.
\end{itemize}

The remainder of this paper is organized as follows: Section \ref{sec:usecases} describes three use cases that are the representatives of real-world distributed systems. Section \ref{sec:riskreq} identifies the challenges and requirements of anomaly detection in the context of real-world distributed systems. Next, Section \ref{sec:wgraph} argues the benefits of moving toward using graphs in distributed systems. Section \ref{sec:graphs} aptly delve into the graph-based approaches and the means of utilizing graphs for anomaly detection. Finally, Section \ref{sec:discuss} discusses and compares these methods and their complexity, the new challenges and requirements that graphs add, and the future directions that this research field can take.
\section{Conceptual Use-cases}
\label{sec:usecases}
In this section, we introduce three conceptual use cases that will be used throughout the paper for better understanding. While these three conceptual use cases are good samples of heterogeneous distributed systems, if necessary, these systems can be segmented into homogeneous subgroups to fit the anomaly detection method's requirements.

We have made sure to cover different levels of abstraction to help with the development of a richer understanding of the concepts. The aim is to clarify that anomalies can occur in all levels of a system, whether it is a small component or it is a large subgroup of the whole system.

\begin{figure}[b]
        \centering
                \includegraphics[width=1\linewidth, height=0.7\linewidth, trim= 1 1 1 1,clip]{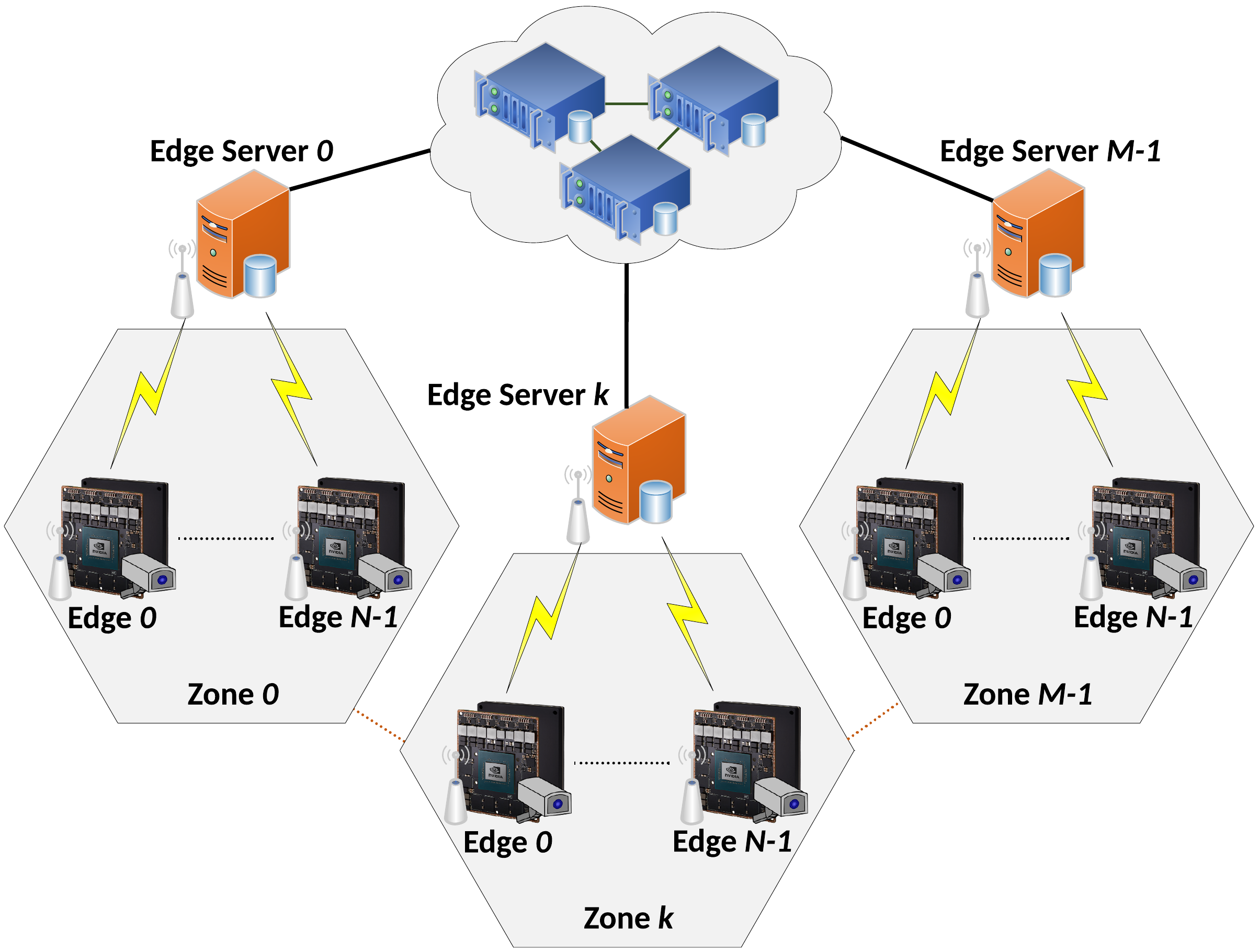}
                \caption{ REVAMP\textsuperscript{2}T \cite{neff2019revamp} setup. This setup can be expanded or shrunk to a new setup. Each edge server only sees the data of its own location and the cloud server is able to see all the locations connected to it. For privacy purposes, a supervised amount of information is transmitted from each location to the cloud server.}
                \label{fig:smart}
\end{figure}
\subsection{Smart Surveillance System}
\label{sec:smartsurv}
Here we use an example from \cite{neff2019revamp}. Real-time Edge Video Analytics for Multi-camera Privacy-aware Pedestrian Tracking or REVAMP\textsuperscript{2}T \cite{neff2019revamp} is designed for tracking pedestrians across multiple cameras. The overview of the model is shown in \figref{fig:smart}. The installed edge nodes at each place of interest consist of a number of edge cameras empowered for analyzing streaming video. These cameras are connected to an edge server that acts as a database that contains information about seen objects. The video can be sent to a number of surveillance monitors with the extracted information if necessary and allowed. Also, the whole information can be sent to a cloud server for further processing of less sensitive information. An anomaly can happen throughout this whole complex distributed heterogeneous system. For example, a camera might get broken, or even there can be a hacker trying to steal the extracted information from the edge server.

This distributed system is completely attributed and dynamic, leading to a dynamic attributed graph representation. Benign changes such as removing or updating the components can happen in this system, which should not be counted as anomalies. Other events such as receiving noisy video from a camera, information mismatch between the edge server and the cloud server, and unwanted access to data at any point of the network can be seen as anomalies. 

\subsection{Sensor Network}
\label{sec:sensor}

\begin{figure}[b]
        \centering
                \includegraphics[width=0.9\linewidth, height=0.9\linewidth]{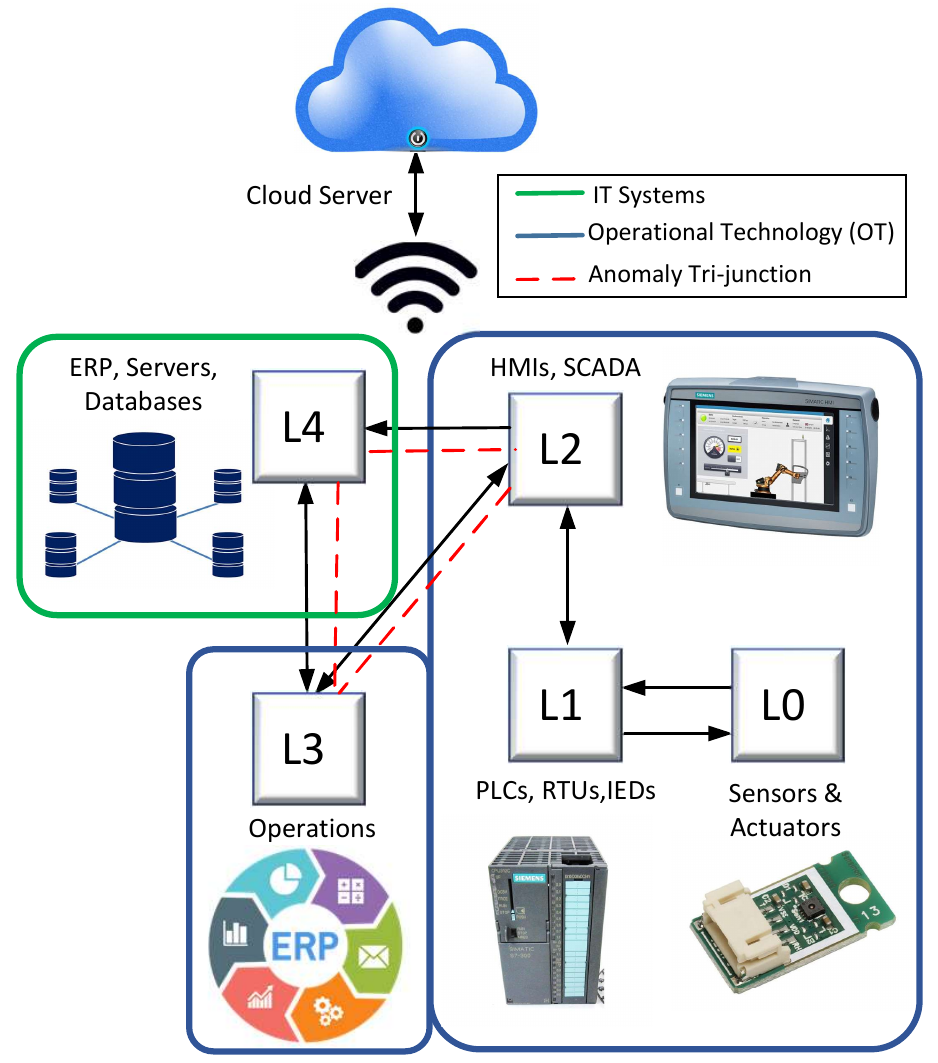}
                \caption{Sensor Network}
                \label{fig:sensor}
\end{figure}

We next look at the conceptual use case of a heterogeneous Sensor Network that illustrates a complex distributed system. To accurately depict the security aspects of such a system, we take inspiration from the Purdue Model for ICS security \cite{purduemodel}. The Purdue Enterprise Reference Architecture was created by mapping the interconnections and inter-dependencies of the high-level components of typical industrial control systems (ICS) to provide guidance on how to defend IT (Information Technology) and OT (Operations Technology) systems against malicious actors. \figref{fig:sensor} shows the model of such a system modified to incorporate all the possible components of a heterogeneous system. Here, Level 0 (L0) through Level 4 (L4) are the functional levels that span two broad zones, namely IT and OT. OT systems can be further classified into two sub-categories which will be discussed as follows.

\begin{itemize}

\item L0: Physical components that build products. Eg. motors, pumps, sensors, valves, etc

\item L1: Systems that monitor and send commands to devices at L0. Eg. PLCs, RTUs, IEDs, etc

\item L2: Overall process controllers. Eg. HMIs, SCADA, etc involves humans managing and controlling the processes

\item L3: Management of production workflows. Eg. batch management, operations management, etc

\item L4: ERP software, databases, email servers and other systems that manage the logistics of the manufacturing operations and provide communications and data storage

\end {itemize}

The anomaly tri-junction barrier (shown in red) serves as a demarcation of communication between the various levels and possible sources of anomalies.
\subsection{Internal Local Area Network}
\label{sec:lan}
An internal Local Area Network (LAN) is also a good sample model of a complex distributed heterogeneous system. The graph representation of the system is attributed, dynamic, and directed. It is dynamic because all the components in the network are subject to change. For example, clients can change their devices, the hardware of the server can change (e.g., adding more memory or upgrading the CPU), the mediums for connections can change, a client can get removed, or the framework of each platform can get updated. 

To emphasize, benign changes to the system must not be counted as an anomaly. We are dealing with a dynamic system that does not have a unique normal state. Thus, capturing anomalies is a more difficult task than when we are dealing with static systems.

\vspace{-10pt}
\section{Challenges}
\label{sec:riskreq}
When it comes to anomaly detection, especially in real-world systems, a number of requirements must be met. In this section, we identify the general requirements and challenges of anomaly detection in real-world distributed systems. We look at these challenges from four aspects, each of which examines anomaly detection from an alternative standpoint: Data, Anomaly Nature, Graph, and Real-world aspects. 

Table \ref{tab:challenge} summarizes all the requirements and the challenges that follow them. In the rest of this section, we discuss each one of them in more detail.

\begin{table*}[htbp]
\renewcommand{\arraystretch}{1.15}

  \centering
  \caption{Challenges of Anomaly Detection.}

    \begin{tabular}{>{\centering\arraybackslash}m{9.855em}||>{\centering\arraybackslash}m{9.855em}|>{\centering\arraybackslash}m{19.855em}|>{\centering\arraybackslash}m{18.5em}}
    
    \multicolumn{1}{>{\centering\arraybackslash}m{8.355em}||}{\textbf{Category}} & \textbf{Challenge} & \textbf{Short Description} & \multicolumn{1}{>{\centering\arraybackslash}m{12.715em}}{\textbf{References}} \bigstrut[b]\\
    \hline
    \hline
    \multicolumn{1}{c||}{\multirow{7}[12]{*}{\textbf{Data and Evaluation}}} & Big Data & Volume, velocity, variety, veracity, and value of big data can be problematic. &\cite{song2016big,katal2013big, 10.1145/3018661.3018676, 10.5555/645924.671334, 10.1145/342009.335437, 10.1145/1321440.1321550, 10.5555/3001460.3001495, 4766873, 5452751} \bigstrut\\
\cline{2-4}          & High-Dimensionality & Many numbers of features for each datapoint can cause sparsity of data which makes anomaly detection more arduous. & \cite{BellmanDynamic, MeaningfulNearest, 10.1111/j.1467-9868.2005.00510.x,10.2307/20441411, 6948273, DataStreamsOutlier, 6228154}
\bigstrut\\
\cline{2-4}          & Heterogeneity & Diversity and variety of data types, formats, and structures in systems. & \cite{iturbe2017towards, erhan2021smart, stiawan2016anomaly, lee2023heterogeneous} \bigstrut\\
\cline{2-4}          & Labeled Data & The number of labeled datasets are alarmingly low for anomaly detection purposes. & \cite{ma2021comprehensive, gornitz2013toward, suthaharan2010labelled, luo2019using, lin2021learning} \bigstrut\\
\cline{2-4}          & Unbalanced Data & Anomalies happen rarely, and training on such unbalanced data makes the model biased. & \cite{frasca2013neural, kalid2020multiple, pajouh2017two, el2020machine}  \bigstrut\\
\cline{2-4}          & Unclean Data & Many proposed models work based on the hypothesis that clean data is available which is not realistic.   & \cite{kang2019robust, evans2018learning, kim2019nlnl, moran2020noisier2noise, zhong2019graph, zaheer2021cleaning} \bigstrut\\
\cline{2-4}          & Metrics and Benchmarks & Lack of Proper Metrics and benchmarks makes the assessment of proposed models difficult and impossible. & \cite{ahmad2017unsupervised, maseer2021benchmarking, emmott2013systematic, banbury2020benchmarking, alinezhad2022adg }\bigstrut\\
    \hline
    \multicolumn{1}{c||}{\multirow{4}[8]{*}{\textbf{Nature of Anomaly}}} & Changing Nature & Novel and new types of anomalies can happen, and anomalies may adapt themselves with anomaly detection algorithms.  & \cite{8944284, 9685997}\bigstrut\\
\cline{2-4}          & Disparate Sources & Anomalies can happen in many different layers and parts of systems. & \cite{9321199, li2019abnormal, ERHAN202164 } \bigstrut\\
\cline{2-4}          & Obscured Anomalies & Outliers can be in disguise, especially the anomalies that are caused by a smart entity.   & \cite{ma2021comprehensive, dou2020enhancing, bhatia2021graphanogan, zheng2021generative} \bigstrut\\
\cline{2-4}          & Noise Resilience & In many cases, noises in the input data can mislead the models and be detected as false positives.   & \cite{liu2020noise, tang2018detection} \bigstrut\\
    \hline
    \multicolumn{1}{c||}{\multirow{5}[9]{*}{\textbf{Real-World}}} & Inference Time & In real-world applications, it is vital to detect anomalies in a timely manner to take the appropriate action.  & \cite{sultani2018real, muharemi2019machine, castellani2020real, 7474197, s150408764} \bigstrut\\
\cline{2-4}          & Privacy & In many applications, the data of the users should be protected. & \cite{liu2020privacy, shen2019online, wang2018graph, yang2020secure, duddu2020quantifying, zhou2020vertically, sajadmanesh2021locally, zhang2021fastgnn, li2021privacy} \bigstrut\\
\cline{2-4}          & Ciphered Data & In many environments, the exchanged data is ciphered. & \cite{bakhshi2021anomaly, pham2021mappgraph, wu2021graph, chen2021learning, protogerou2021graph} \bigstrut\\
\cline{2-4}          & Dynamic Systems & Real-world systems often change over the time and have dynamic nature. & \cite{liu2021anomaly, zhou2021anomaly, ma2021comprehensive, zheng2019addgraph, pourhabibi2020fraud, salehi2018survey, ding2019interactive, xue2022dynamic} \bigstrut\\
\cline{2-4}          & Interpretability & In real-world setups, the cause of the anomaly is important. & \cite{soldani2022anomaly, deng2021graph, wang2021groot} \bigstrut\\
\cline{2-4}          & Domain Shift & The model trained on one domain, cannot generalize even to a slightly different domain.  & \cite{ding2021cross,  ganin2015unsupervised, hoffman2014lsda, collobert2011natural, li2012literature, sogaard2013semi, zhang2020collaborative} \bigstrut[t]\\
    \end{tabular}%
  \label{tab:challenge}%
\vspace{-10pt}
\end{table*}%

\vspace{-10pt}
\subsection{Data and Evaluation Aspect}
\textbf{1. Big Data:} When it comes to anomaly detection with learning approaches, the data becomes an immediate challenge. In many applications, the amount of data that has to be processed is immense, pushing the problem into the big data category. Big data meaning is best described through the 5Vs of big data: Volume, Velocity, Variety, Veracity, and Value, each exhibiting a series of challenges. For anomaly detection, especially in distributed systems, a huge amount of data (Volume) is being generated and processed at a very high speed (Velocity), containing multiple data types and formats coming from several sources (Variety), with uncertainty about the data and its quality (Veracity), all or most of it important, and must be considered for extracting a worthy output (Value).

\textbf{2. High Dimensionality:} This is another challenge that arises from the data. In many applications, the data that should be processed has a large number of features (dimensions). Huge dimensions of the dataset can cause numerous challenges in many applications. However, these challenges are exacerbated in the context of anomaly detection. The phrase “curse of dimensionality” generally refers to the problems that arise when the number of dimensions increases. A dataset is high-dimensional when we can see the curse of dimensionality. The growth of dimensions will increase the size of the data accordingly and causes sparsity which eventually results in the data points having relatively the same distance from each other. As a result, it would be more difficult to detect hidden anomalies in high-dimensional space. This issue is still a controversial topic in scientific society.

\textbf{3. Heterogeneity:} Distributed systems, specially in real-world scenarios, normally include various types of entities, each with their own set of characteristics and behaviors. Consequently the relation between these components is also specific based on the type of the component. The data produced from such systems are heterogeneous, and many methods are not able to handle complex variable heterogeneous data. Thus, heterogeneity is another challenge for the task of anomaly detection. Heterogeneous systems often have large and complex data structures, with a high degree of interconnectivity and variability.

\textbf{4. Labeled Data:} In data-driven approaches, the quality of the data has a vital role in the outcome of the model. When it comes to anomaly detection, the number of labeled datasets is alarmingly low. This fact encourages the use of unsupervised or semi-supervised approaches. Thanks to new technologies and advancements in devices, a huge amount of data is available. However, in most cases, this data does not incorporate information regarding anomalous behavior. On the other hand, it is both expensive and time-consuming to obtain high-quality labels for available data. In many cases, anomalies must be discovered and labeled by experts who have field-specific knowledge, making the task harder and more demanding. Also, we should consider the noises added to the dataset by imprecise labeling. Undetected anomalies can have a huge cost in critical situations of real-world applications. Thus, using supervised learning is somewhat problematic in the context of anomaly detection. However, several approaches try to overcome this challenge by taking advantage of synthetic labeled datasets, but many other approaches have moved toward using semi-supervised, unsupervised, and self-supervised algorithms.

\textbf{5. Unbalanced Data:} Another point to be mentioned is that anomalies are rare. Supervised learning approaches need to see enough examples while training to learn and understand anomalies. However, the number of anomalies is usually very subtle, causing the dataset to become unbalanced. Anomalies are out-of-the-ordinary incidents, and it is not rational to expect enough samples of each type of anomaly to be available in the dataset. This challenge requires the direction of solving the problem to adapt to it. The abundance of normal data points may result in neglecting the detection of anomalous data points which are valuable in anomaly detection. This problem arises from the fact that many machine learning algorithms are based on the hypothesis that classes of data have the same distributions which is not realistic in the case of anomaly detection.

\textbf{6. Unclean Data:} Many approaches, such as anomaly detection with autoencoders (discussed in Section \ref{sec:graphs} need datasets that only contain normal datapoints for training. These approaches learn the normal state and decide whether an anomaly has happened or not based on the divergence from features of normal datapoints. However, as discussed, anomalies can be rare and very subtle. There is a very high chance that an anomalous datapoint slips through the fingers of the responsible individual or team, or it might not even be considered an anomaly until later. As a result, often datasets include undesired outliers that make the dataset unclean. This unclean data can confuse the model about the features and representation of the normal datapoints and deteriorate the accuracy of the model. 

\textbf{7. Metrics and Benchmarks:} On top of all the challenges, the most important one seems to be the fact that there is no unified metric or benchmark, where models can assess themselves and compare the results. Various kinds of metrics have been introduced and we discuss a number of important ones in Section \ref{sec:metrics}, but almost none of the works assess those metrics on a single unique domain, to make the comparison between models possible. 

\subsection{Anomaly Nature}
\textbf{1. Changing Nature:} Anomalies have a specific nature and a collection of characteristics that lead to a particular set of challenges. The first and most important challenge is the changing nature of anomalies. Anomalies are not predefined static occurrences. They can transpire in different shapes and formats, and not all of them can be considered prior to the occurrence of an anomaly. There can always be novel anomalies that have never been considered. They can adapt to anomaly detection mechanisms and evolve to pass through the defenses of anomaly detection algorithms. Because of this characteristic, most algorithms tend to skew toward online learning, unsupervised learning, or semi-supervised learning, where there are no predefined anomalies, and the algorithm learns based on what it sees at the moment on its own. In other words, most of the algorithms make an effort to eliminate external artificial biases and proceed toward generic solutions. This instigates the use of online learning, where the algorithm is constantly learning about the behavior of the system.

\textbf{2. Disparate Sources}: Distributed systems typically incorporate multiple layers (resources), resulting in a complex and intricate system as a whole. The role of each of these layers, disregarding its majority or minority can not be neglected. Anomalies can penetrate each and every one of those components, making it very difficult to track them. Especially in large facilities, the number of components grows exponentially, and all of them are potential points of anomaly. This disparate source of anomalies is one of the challenging complications that is troubling anomaly detection algorithms.

\textbf{3. Obscured Anomalies:} In addition to these challenges, a considerable number of anomalies, predominantly the ones that came into existence because of a smart entity, do not like to be found. They might be disguised, or hidden. For example, a cyber-attack tries to be as masked as possible. On the other hand, there are anomalies that are not concealed and are just the result of a malfunctioning component in the system. Being conscious of both of these aspects and prepared for obscured anomalies is another challenging task on the way to generic anomaly detection.

\textbf{4. Noise Resilience: }Systems are subject to experiencing a profuse number of internal or external noises. These noises are not anomalies until they pass a certain threshold. However, they are still an irregularity in the system. The anomaly detection algorithms must be capable of distinguishing between noises and anomalies. If not, they might raise too many false positive alarms, resulting in supererogatory attention and costs. These noises can have a wide variety of origins. Some of them are back-breaking to discover, and it is best to introduce a means of filtering those noises out inside the algorithm itself, instead of putting an effort to find the source of the noise.
\vspace{-10pt}
\subsection{Real-world Challenges}
\textbf{1. Inference Time:} When it comes to using anomaly detection, time is of the essence, especially in real-world applications. It is expected, time-wise, that the detection of the anomaly does not sit far from the actual happening of the anomalous behavior. Particularly, in a critical environment such as healthcare applications, a long latency in detecting a malfunctioning device might lead to catastrophic results, and it is impermissible. Thus, necessary actions must be taken to avoid such a latency. However, anomaly detection techniques based on deep learning are often a heavy task, requiring a colossal amount of computational power.

\textbf{2. Privacy:} Privacy is another important hindrance that must be considered in a real-world environment. Anomaly detection algorithms are involved with all the data circulation and the state of the system to the furthest extent. They know everything about a system, and if that kind of information leaks out of the environment, it can lead to cataclysmic undesirable results. Hence, an anomaly detection algorithm must be obliged to preserve privacy and be as secure as possible if it desires to be deployed in a real-world environment.

\textbf{3. Ciphered Data:} The fact that anomaly detection algorithms need to be aware of everything that is going on related to a system brings about another challenge. Many systems such as financial systems, social media, or facilities like power generation and distribution networks do not intercommunicate in plain, raw data. They encode the data and then transmit it. This ciphered data usually is not related to the original data in an obvious manner and that is the whole purpose of encoding the data. Hence, algorithms face yet another arduous challenge. For example, wireless communication between a sensor and the server contains an encoder at the edge transmitter, ciphering the data for privacy and security reasons, and a decoder at the receptor that deciphers the data for further processing. When an algorithm tries to discover anomalies on the medium (here the wireless connector) it cannot access the raw data before the transmission stage.

\textbf{4. Dynamic Systems:} Furthermore, static systems are extremely rare in a real-world environment. Real-world distributed systems are extremely dynamic. Components get removed, added, changed, or updated. These benign transformations must not be counted as anomalous behavior. The anomaly detection algorithms might consider these transformations as anomalies. If so, the number of false-positive alarms might rise to a point that makes the algorithm non-functional and impractical for real-world adoption. On the other hand, the anomaly detection model should be flexible enough to learn about the updated features and structure of the network to fully incorporate existing information. Online learning is a tool for overcoming this challenge, however, this is still a very hot topic in research communities.

\textbf{5. Interpretability:} Another important challenge is the interpretability of the anomalies. Particularly, in real-world deployments, the tendency to see the actual type of anomaly and not just the detection is more desired. One reason behind this is that recognizing the source of the anomaly, realizing the actual type of the anomaly (e.g., cyber attacks on sensor 1A), and informing the responsible individuals is very important. Finding the root of the anomaly is important for maintaining the availability of the system.

\textbf{6. Domain Shift:} Last but not least, anomaly detection approaches become handy, when you are able to generalize over different domains. However, since we are dealing with learning-based approaches, this domain shift comes with a huge cost. Neural Networks learn about the environment and adapt to it, making it difficult to switch to another domain. This fact discourages the use of supervised learning and also pushes novel techniques toward online learning. This challenge is shared between graph-based anomaly detection and other approaches.
\vspace{-10pt}
\section{Motivation: Why Graph-based?}
\label{sec:wgraph}

\begin{table*}[htbp]
  \centering
  \caption{Summary of Reviewed Approaches and Their Capabilities} 

    \begin{tabular}{P{10.855em}||cc|c|c|c|c}
    \multicolumn{1}{c||}{\textbf{Approach}} & 
    \multicolumn{2}{c|}{\textbf{Model}} & 
    \multicolumn{1}{P{5em}|}{\textbf{Attributed Graphs}} & 
    \multicolumn{1}{P{5.0em}|}{\textbf{Dynamic Graphs}} & 
    \multicolumn{1}{P{6.5em}|}{\textbf{Learning Adaptability}} & 
    \multicolumn{1}{P{5.5em}}{\textbf{Scalability}}
    \bigstrut[b]\\
    \hline
    \multicolumn{1}{c||}{\multirow{2}[7]{*}{\textbf{Federated Learning}}} & \multicolumn{2}{c|}{DIoT\cite{2019diot}} & \xmark    & \cmark    & S  &  \bigstrut\\
\cline{2-7}    \multicolumn{1}{c||}{} & \multicolumn{2}{c|}{VFL\cite{vfl2021}} & \xmark    & \cmark    & U    &      \bigstrut\\
    \hline
    \multicolumn{1}{c||}{\multirow{9}[18]{*}{\textbf{Autoencoders}}} & \multicolumn{2}{c|}{\cite{borghesi2019anomaly}} & \xmark    & \xmark& U    & \cmark  \bigstrut\\
\cline{2-7}    \multicolumn{1}{c||}{} & \multicolumn{2}{c|}{ \cite{8363930}} & \xmark   & \xmark   & U  &    \bigstrut\\
\cline{2-7}    \multicolumn{1}{c||}{} & \multicolumn{2}{c|}{ \cite{8532358}} & \xmark   & \xmark   & U   &   \bigstrut\\
\cline{2-7}    \multicolumn{1}{c||}{} & \multicolumn{2}{P{17.145em}|}{Robust Deep Autoencoder (RDA) \cite{10.1145/3097983.3098052}} & \xmark& \xmark& U &    \bigstrut\\
\cline{2-7}    \multicolumn{1}{c||}{} & \multicolumn{2}{P{17.145em}|}{Iterative Training Set Refinement (ITSR)  \cite{10.1007/978-3-030-46150-8_13}} & \xmark   & \xmark   & U    &  \bigstrut\\
\cline{2-7}    \multicolumn{1}{c||}{} & \multicolumn{2}{c|}{\cite{CGN_particle}} & \cmark   & \cmark   & U     &  \bigstrut\\
\cline{2-7}    \multicolumn{1}{c||}{} & \multicolumn{2}{c|}{DONE \cite{bandyopadhyay2020outlier}} & \cmark   & \xmark   & U      & \cmark \bigstrut\\
\cline{2-7}    \multicolumn{1}{c||}{} & \multicolumn{2}{c|}{AdONE \cite{bandyopadhyay2020outlier}} & \cmark   & \xmark   & U     & \cmark \bigstrut\\
\cline{2-7}    \multicolumn{1}{c||}{} & \multicolumn{2}{c|}{DeepSphere \cite{teng2018deep}} & \xmark   & \cmark   & U    &  \bigstrut\\
    \hline
    \multicolumn{1}{c||}{\multirow{7}[14]{*}{\textbf{Graph Embedding}}} & \multicolumn{1}{c|}{\multirow{5}[10]{*}{Shallow Encoders}} & DeepWalk \cite{10.1145/2623330.2623732} & \xmark   & \xmark   & U, O   & \cmark\bigstrut\\
\cline{3-7}    \multicolumn{1}{c||}{} & \multicolumn{1}{c|}{} & Node2Vec \cite{10.1145/2939672.2939754} & \xmark   & \xmark   & SM   & \cmark\bigstrut\\
\cline{3-7}    \multicolumn{1}{c||}{} & \multicolumn{1}{c|}{} & LINE \cite{10.1145/2736277.2741093} & \xmark   & \xmark   & -   &   \cmark\bigstrut\\
\cline{3-7}    \multicolumn{1}{c||}{} & \multicolumn{1}{c|}{} & TADW \cite{yang2015network} & \xmark   & \xmark   & -   &  \bigstrut\\
\cline{3-7}    \multicolumn{1}{c||}{} & \multicolumn{1}{c|}{} & NetWalk \cite{10.1145/3219819.3220024} & \xmark   & \cmark   & U, O   & \bigstrut\\
\cline{2-7}    \multicolumn{1}{c||}{} & \multicolumn{1}{c|}{\multirow{2}[4]{*}{Deep Encoders}} & \multicolumn{1}{P{9.5em}|}{Graph Deviation Network (GDN) \cite{deng2021graph}} & \cmark   & \xmark   & U   &     \bigstrut\\
\cline{3-7}    \multicolumn{1}{c||}{} & \multicolumn{1}{c|}{} & AddGraph \cite{ijcai2019-614} & \cmark   & \cmark   & SM    &  \bigstrut\\
    \hline
    \multicolumn{1}{c||}{\multirow{2}[7]{*}{\textbf{Graph Transformers}}} & \multicolumn{2}{c|}{\cite{feng2021heterogeneity}} & \xmark    & \cmark    & S  & \bigstrut\\
\cline{2-7}    \multicolumn{1}{c||}{} & \multicolumn{2}{c|}{TADDY\cite{9599560}} & \xmark    & \cmark    & U    &      \bigstrut\\
\cline{2-7}    \multicolumn{1}{c||}{} & \multicolumn{2}{c|}{GTA \cite{yu2015multi}} & \xmark    &   \cmark  & U    &   \bigstrut\\
    \hline
    \multicolumn{1}{c||}{\textbf{Graph Signal Processing}} & \multicolumn{2}{c|}{\cite{spectralgsp}} & \cmark  & \cmark    & U  & \bigstrut\\
    \hline
    \multicolumn{1}{c||}{\multirow{2}[7]{*}{\textbf{Graph Contrastive Learning}}} & \multicolumn{2}{c|}{GCCAD \cite{chen2021gccad}} & \cmark    & \cmark    & U, SF    &  \bigstrut\\
\cline{2-7}    \multicolumn{1}{c||}{} & \multicolumn{2}{c|}{CoLA \cite{liu2021anomaly}} & \cmark    & \cmark    & SF    &   \cmark \bigstrut\\
\cline{2-7}    \multicolumn{1}{c||}{} & \multicolumn{2}{c|}{SL-GAD \cite{9568697}} & \cmark    & \cmark    & SF  &  \cmark \bigstrut\\
\multicolumn{7}{l}{\scriptsize \textbf{S}: Supervised Learning, \textbf{SM}: Semi-supervised Learning, \textbf{U}: Unsupervised Learning, \textbf{SF}: Self-supervised Learning, \textbf{O}: Online Learning}
    \end{tabular}%
  \label{algorithms}
\vspace{-10pt}
\end{table*}%

The use of graphs for anomaly detection is a very recent topic of interest in the machine learning and deep learning community. In this section, we highlight the benefits of graphs in the context of anomaly detection.

\textbf{Ability to Represent Complex Dependencies:} Graphs are widely used for modeling and analyzing complex systems with non-Euclidean data and intricate relationships among system components. This characteristic is particularly useful for anomaly detection problems since anomalies in nature can arise from complex interactions between variables that are hard to capture using traditional models.

\textbf{Flexibility:} Complex systems consist of many different components each with its own set of characteristics. Graph-based models are highly flexible and can be adapted to a wide range of structures and data types. Graphs suit well for both structured (symbols, images, grid-based data) and unstructured (knowledge graphs, social network data, distributed systems, citation networks, network traffic) that can be easily represented by regular and/or irregular graph structures.

\textbf{Scalibilty}: Anomaly detection often deals with high-dimensional data with a large number of features. Due to this fact, it is hard to capture the underlying patterns and distribution of data for finding outliers. Graphs are useful for capturing the underlying structure of the input data and its relationships in a more compact and interpretable format. This can help with more efficient processing of the data and helps with scalability.

\textbf{Robustness:} Graphs are capable of incorporating relationships and the holistic structure of the data which is particularly beneficial in anomaly detection. Graph-based algorithms analyze each datapoint by taking into account the context related to them which improves the robustness and helps reduce the number of false negatives. This is due to the fact that often outliers do not fit into the global patterns of the system.

\textbf{Interpretability and Visualization:} Graph-based models can provide more insights into the relationships between nodes of the system, making them more interpretable than many traditional models. This can be particularly useful in understanding the underlying causes of anomalies and identifying potential areas for improvement and vulnerabilities of the system.
\vspace{-10pt}
\section{Graph-based Tools and Techniques for Anomaly Detection}
\label{sec:graphs}

In this section, we aim to introduce and compare techniques that combined with graphs have a great potential for solving the problem of anomaly detection in distributed systems. The summary of all reviewed works and their capabilities can be seen in Table \ref{algorithms}.

Many of the surveyed methods are not able to handle heterogeneity or dynamic behavior that might be present in the systems. Approaches that are capable of managing attributed graphs are better suited for heterogeneous systems. It is noteworthy that the inability of certain methods to handle attributed graphs does not mean they can not be adopted in heterogeneous systems. Two viable strategies exist for their utilization. Firstly, large systems can be subdivided into smaller homogeneous sub-systems. Secondly, a simplified version of data can be employed to accommodate the limitations of such methods.

\vspace{-10pt}
\subsection{Federated Learning}

In the context of anomaly detection in distributed systems and when dealing with big data, while graphs can offer the ability to capture complex dependencies and handle scalability and robustness, Federated Learning (FL) can be a solution for improved privacy, model generalization, and more efficient learning. The combination of the two seems to be perfectly suited for anomaly detection in distributed systems, and it is a great area to be explored by scholars and researchers.

Federated Learning is a technology that has recently emerged as an alternative to centralized systems. It focuses on collaboration while preserving the security aspect of client data information used for training machine learning algorithms \cite{aledhari,bonawitz2019,chen2019federated,kon2017federated,wang2020federated}. It also greatly reduces communication overhead \cite{commeff}. FL employs collaborative and experiential learning by training without the need to transfer data over to a centralized location. This feature has recently led to a rise in the applicability of FL in a variety of applications, ranging from medical to IoT, transportation, defense, and mobile apps.

\begin{figure*}[h]
        \centering
               \includegraphics[width=0.9\linewidth, height=0.4\linewidth]{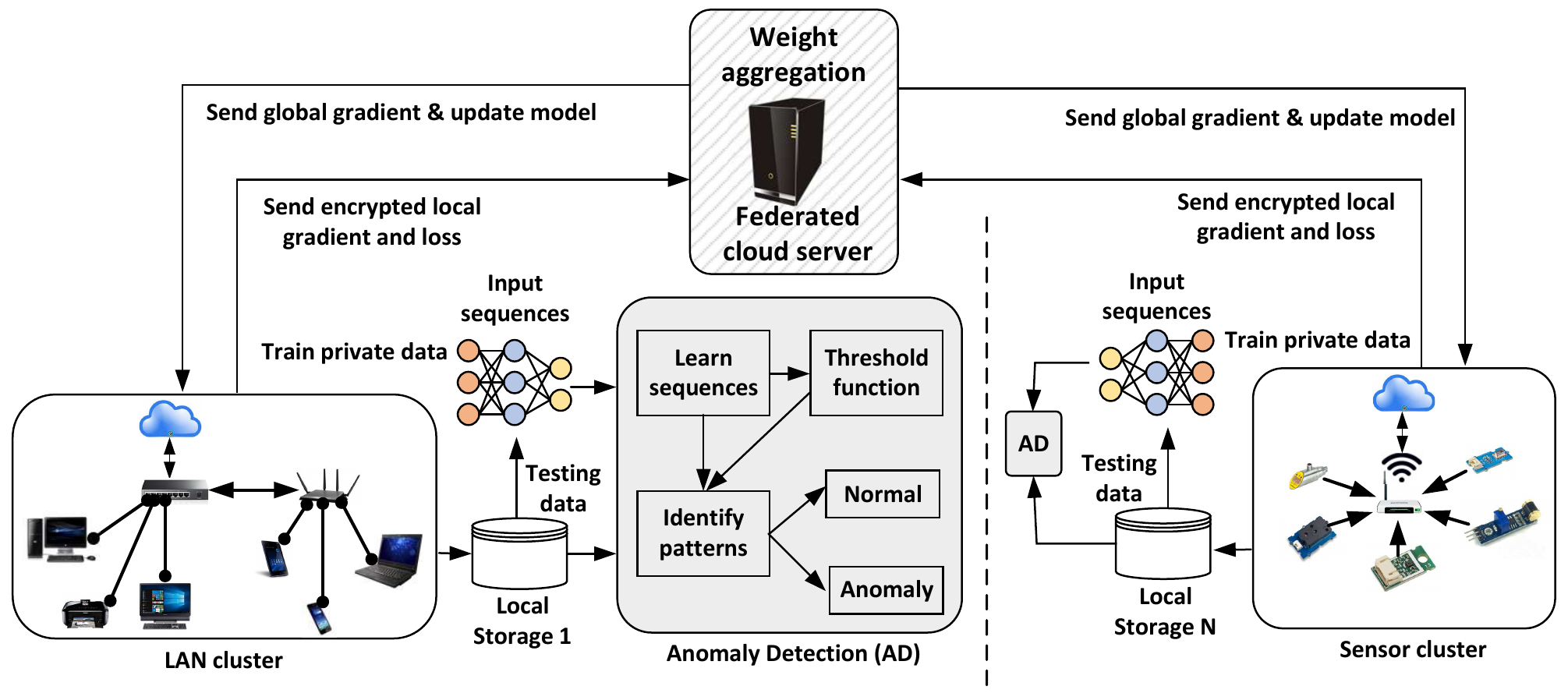}
                \vspace{-15pt}
                \caption{Federated learning for anomaly detection applied on two distributed clusters viz. LAN and Sensor. Here N is the number of clusters connected to the Federated cloud server. Each network has its own testing and training dataset, local storage, and loss function. The server receives individual weights from all the network's local models and then aggregates them based on their useable graph attributes while sending back the global weights.}
                                \vspace{-10pt}
                \label{fig:federated}
\end{figure*}

Anomaly detection and prediction of static and dynamic time series data have been a popular topic, especially for IoT-based data \cite{chandra41,leakdetection,chen7}. These approaches look at multi-sensor systems as a collection of centralized sensors where abnormal sensor behavior is detected by a central model that runs on the server. Such systems are prone to failures, require longer access times, and are vulnerable to malicious invasion attacks leading to a potential data breach from clients to the server \cite{Goodhue}. \cite{Raedafed} looks at the data collected by IoT sensors for energy-efficient applications like HVAC in smart buildings and proposes a federated stacked Long Short-time Memory model (LSTM) on time series data generated by IoT sensors for classification and regression tasks. \cite{2019diot}, on the other hand, proposes a self-learning distributed system for security monitoring.

While such approaches deal with Autonomy (concerns association and communication requirements) nature of systems, they fall short of the benefits of graph-based relational learning. Federated Graph Learning is a fairly recent topic. \cite{zhang2021federated} introduces Federated Graph Learning (FGL) as a seminal paper discussing the definition and challenges of FGL. It further categorizes FGL into four distinct learning types. Graph edges and nodes in each sub-graph heavily overlap each other and therefore play a pivotal role in extracting the features and adding to contextual and relational learning benefits. A sub-classification of this work was introduced in \cite{vfl2021} as a complete methodology to apply the Vertical Federated Learning (VFL) algorithm for graph convolutional networks. The approach, along with Homomorphic Encryption (HE), was developed to ensure privacy while maintaining accuracy.

A general structure of a complex distributed system can be easily depicted using the Federated Learning framework (Fig \ref{fig:federated}). Here we show two different network clusters. In this framework, the models are trained at the device level or client side, where they are brought over to the data sources or devices for training and prediction. The updated values are sent back to the federated cloud server for aggregation. One consolidated model gets transferred back to the devices to enable tracking and redistribution of each model to various devices. During the training phase, the input is reconstructed in the output until reconstruction error is minimized, which calculates the threshold value. This threshold value decides whether the observed patterns are anomalous or not.

Table \ref{algorithms} summarizes the strong points and weaknesses of two discussed works DIoT \cite{2019diot} and VFL \cite{vfl2021}. Both of these works are unable to process the additional information provided by attributed graphs which in real-world scenarios are common. It should be mentioned that for supervised models such as DIot \cite{2019diot} a crucial requirement is available labeled data that is hard to provide. Based on mentioned shortcomings, although FGL addresses data diversity, data security and real-time continuous learning it is not mature enough yet for real-world applications.

\begin{figure*}[t]
        \centering

                \includegraphics[width=1\linewidth, height=0.25\linewidth, trim= 2 2 2 2,clip]{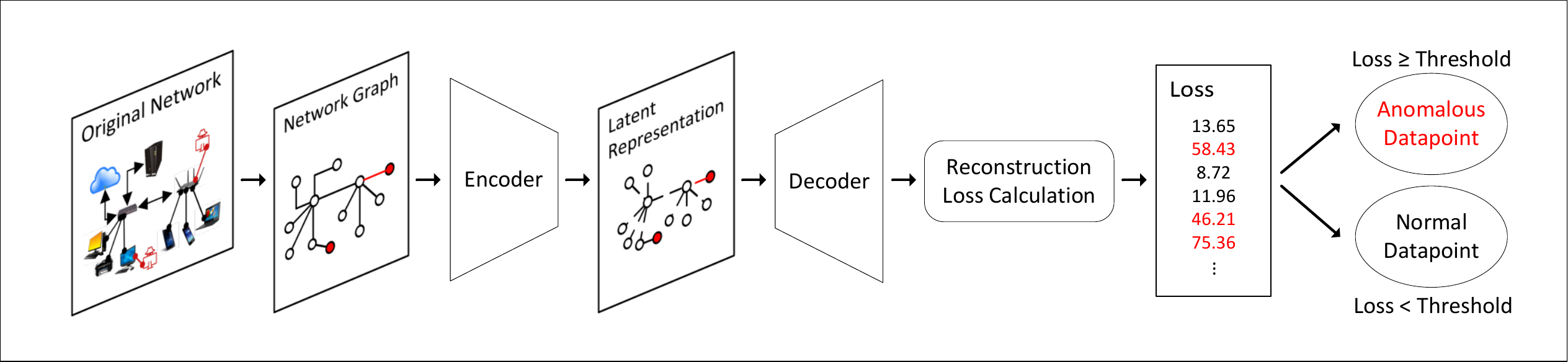}
                            \vspace{-30pt}
                \caption{Graph autoencoder network for anomaly detection in the LAN Cluster example introduced in Section \ref{sec:lan}. Two malicious users (equal to anomalous nodes in the network graph) have connected to the network. The network graph is fed to the autoencoder and the reconstruction loss for nodes and edges will be calculated. In this simple case, just by comparing the reconstruction loss to a constant threshold, the anomalies can be detected, but these processes can be more advanced and precise.}
                \vspace{-10pt}
                \label{fig:autoencoder}
\end{figure*}

\subsection{Autoencoders}
Anomaly Detection is often considered an open-set problem, where it is very unlikely to have complete knowledge of all types of possible anomalies. As a result, researchers have focused on developing semi-supervised or unsupervised powerful neural network models such as Autoencoders. The combination of autoencoders with graphs has become a popular topic as it helps with overcoming the challenges of anomaly detection, specifically in the context of distributed systems.

Autoencoders are typically made of two main modules; encoders and decoders. The encoder part is responsible for mapping the input space to a bottleneck latent space, and the decoder reconstructs the original input based on the latent representation. In order to find the best possible latent representation, they try to minimize the reconstruction error of the original input. Thus, the network will learn to preserve the most informative parts of the input features in the latent representation. They were traditionally used for dimension reduction prior to feeding the data to the main network, but nowadays, they have vast applications in information retrieval, image processing, and anomaly detection. 

In the context of anomaly detection, autoencoders can be used as a tool for calculating anomaly scores. After training the autoencoder with data that does not contain anomalous points, if any outlier data points are fed to the network, it will perform poorly. The reconstruction loss of that data point will be larger compared to normal ones since the network is not familiar with this type of data. As a result, the reconstruction error can be used as a measure of deviation from normal data points. This approach has been used for detecting anomalies in high-performance computing systems \cite{borghesi2019anomaly} and has shown promising results. Convolutional Autoencoders can improve the parameter efficiency and training time since they have shared parameters. Works such as \cite{8363930, 8532358}
have made use of convolutional autoencoders for anomaly detection and achieved significant improvements. DeepSphere \cite{teng2018deep} proposed a method for detecting anomalous snapshots in a dynamic network. This work adopts an LSTM autoencoder with an attention mechanism. In the constructed hidden space, DeepSphere \cite{teng2018deep} learns a spherically shaped boundary around the encoded normal representations. As a result, the encoded representation of an unseen anomalous snapshot of the network will fall outside the hyper-sphere and is detected as an anomaly. 

All the aforementioned methods are developed based on the hypothesis that clean data (data that does not contain anomalous points) is available for training, but in many real-world cases, we do not have enough data points that satisfy these constrain. Thus, the reconstruction error for anomalous points will be lower, and the accuracy of finding outliers will degrade. To solve this issue, Robust Deep Autoencoder (RDA) \cite{10.1145/3097983.3098052} inspired by Robust Principal Component Analysis \cite{10.1145/1970392.1970395, Donoho2006ForML, 6400214} uses a filter layer that separates anomalous data points of input data. By removing these noisy and anomalous data points, the network will be able to better reconstruct the normal data points. On top of RDA \cite{10.1145/3097983.3098052}, they add an anomaly detection algorithm to test the effectiveness of the proposed method. Iterative Training Set Refinement (ITSR) \cite{10.1007/978-3-030-46150-8_13} adapts adversarial autoencoder network \cite{44904} architecture to add a prior distribution to constructed latent representation and places anomalies to the regions with lower likelihood. By this means, the model will be robust against noises and anomalies in the training data. DONE \cite{bandyopadhyay2020outlier} is another network proposed for detecting anomalous nodes in attributed graphs. DONE makes use of two parallel autoencoders; one for encoding link structure and another one for attributes of nodes. These autoencoders are trained to preserve proximity and homophily in the network. The proposed loss function is designed to minimize the contribution of outliers, and by minimizing the loss equation, the anomaly scores for each node are calculated. Finally, the top k nodes are reported as anomalous points. They also propose AdONE \cite{bandyopadhyay2020outlier} that makes use of adversarial learning to be able to construct an outlier-resistant network embedding.

Traditional autoencoder networks and convolutional autoencoders are limited to a fixed input length which makes them unsuitable for detecting anomalies in dynamic networks. Also, they are limited to Euclidean data; thus, they are unable to model complex relationships that may occur in a dynamic heterogeneous network. Graph Neural Networks (GNN) can overcome these issues and generalize the network. They can also incorporate multi-dimensional edge and node attributes. Convolutional graph autoencoders have been used in \cite{CGN_particle} for detecting anomalies. This model uses both node features and edge features. Also, the decoder module has two separate branches for node reconstruction and edge reconstruction. The final reconstruction loss is the combination of edge reconstruction loss and node reconstruction loss and is used as a measure for finding anomalies. 

Fig \ref{fig:autoencoder} shows the workflow for detecting anomalies using autoencoder networks in the LAN Cluster example mentioned in Section \ref{sec:lan} . In the case of anomaly detection in a complex system, such as the three conceptual models that we introduced earlier, after training the model with normal data, the graph of the network is fed to the encoder module to be transformed into the latent space. In this case, since we are dealing with a heterogeneous network, the encoder and decoder modules should be able to incorporate the edge and node information and aggregate them to get the most of the available information. Also, at any time, the graph of the network can change. Thus, it is vital to employ a network architecture that can handle dynamic graphs, such as Graph Neural Networks. The decoder will try to reconstruct the original graph of the network from the latent representation. Anomalous edges, nodes, or sub-graphs of the network will have a high reconstruction error. In the most basic approach, only a fixed threshold can be used to decide whether the edge, node, or sub-graph should be considered anomalous or not. 

For more clear scrutiny, Table \ref{algorithms} shows the different features of reviewed works. As mentioned before, one important characteristic of the autoencoder structure is that it is trained in an unsupervised manner that directly solves the problem of the availability of labeled data. Also, the changing nature of anomalies will not be a problem in these types of models since they will learn the normal behavior of the system and anything different from that will be counted as an outlier.
 
\begin{figure*}[!h]
        \centering
                \includegraphics[width=.8\linewidth, trim= 20 25 20 15,clip]{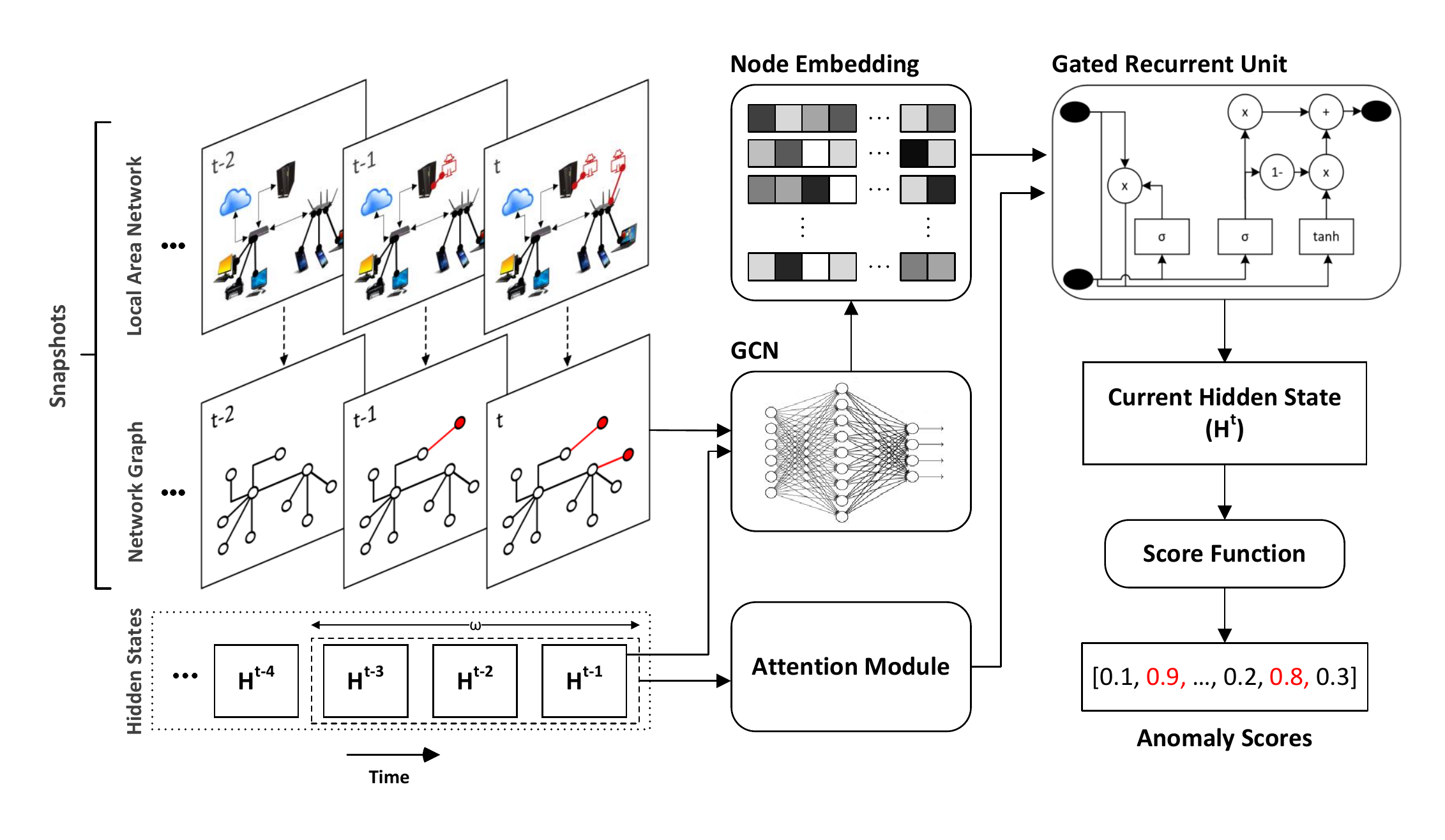}
                \vspace{-10pt}
                \caption{This figure shows the AddGraph model applied to the LAN Cluster conceptual use case discussed in \ref{sec:lan}. In the first step, Graph Convolutional Network (GCN) takes the graph snapshot at time step t and combines it with the hidden state of the previous time step to construct the node embedding. GRU and attention module combine the long-term and short-term states to generate the current hidden state. In the last step, the scoring function assigns an anomaly score to edges based on the nodes connected to them.}
                 \vspace{-10pt}
                \label{fig:addgraph}
\end{figure*}
\vspace{-10pt}
\subsection{Graph Embedding}
The first obstacle that we face in many tasks like anomaly detection in large networks is finding a way to map the data hidden in the network graph into a low-dimensional space. To do so, there are many different approaches. In general, we can separate these methods into two different classes. The first type of encoder is “Shallow Encoders” which will transform each node of a graph into exactly one vector in the latent space. On the other hand, there are methods called “Deep Encoders” which get the use of more complicated networks and are able to generate more complex embedding compared to “Shallow Encoders.” In the following, we will see details of each of these methods and their strengths and weaknesses. After obtaining the latent representations, an anomaly detection algorithm can be used on top of the embedding network in order to detect outliers and anomalies.
\vspace{-5pt}
\subsubsection{Shallow Embedding}
Shallow Embedding methods try to find a unique latent representation for each vertex of a graph. The main difference between different approaches to shallow embedding is in the definition of similarity function. The similarity function basically describes how relationships in the latent space are mapped to the original input space. DeepWalk \cite{10.1145/2623330.2623732} is a random walk approach for node embedding that tries to find the best embedding that preserves similarity. Each random walk is an unbiased sequence of nodes with a fixed length. Authors claim that random walks can be treated as sentences of a text since the frequency of occurrence of the nodes in random walks follows the power law. DeepWalk \cite{10.1145/2623330.2623732} tries to find the feature representation for each node such that it maximizes the likelihood of visiting nodes seen in the random walks starting from that particular node. Another method to be mentioned is Node2Vec \cite{10.1145/2939672.2939754}. This algorithm makes it possible to have more flexible random walks in order to obtain richer latent representations. It combines breath-first sampling and depth-first sampling together to introduce a flexible biased sampling strategy that allows local and global views of the network. 

Similar approaches such as LINE \cite{10.1145/2736277.2741093}, and TADW \cite{yang2015network} have been proposed. One might think of these algorithms as finding a simple look-up table for assigning each node in the original graph to a latent representation. As a result, these kinds of methods have limitations when we want to apply them to large heterogeneous networks. The first limitation comes from the fact that each node has to have its own unique embedding and there are no shared parameters in these kinds of networks. As a result, when the number of nodes grows, the number of parameters will grow respectively, and we will need V$\times$D number of parameters where D is the dimension of latent space. Also, they are unable to generalize to unseen nodes, and we cannot use them in dynamic networks. NetWalk \cite{10.1145/3219819.3220024} solves the problem with the changing networks and dynamically updates the representations as the network evolves. This model tries to satisfy two constraints: clique constraints which minimize the pairwise distances between representations of vertices in each random walk to preserve locality and autoencoder constrain, which serve as a global constraint and minimize the reconstruction error of input using the output embedding. Still, there are more issues with these types of graph embedding algorithms; another problem is that in complex heterogeneous systems, each node may have its own specific features, but “Shallow Encoders” are unable to make use of these node-specific features.                                      
\begin{figure*}[!h]
        \centering
                \includegraphics[width=\linewidth, trim= 18 18 23.5 16,clip]{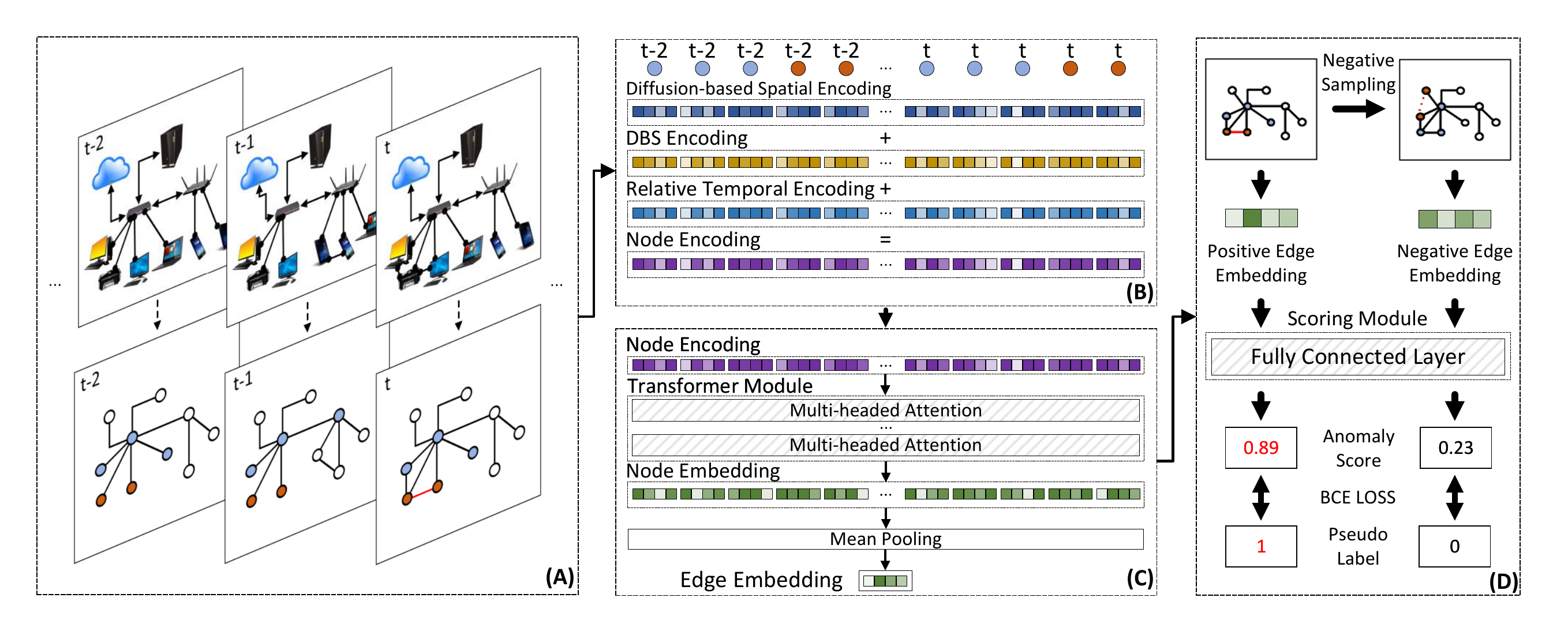}
                \caption{The figure shows the frame work of TADDY \cite{9599560}. (A) shows the edge-based substructure sampling module which chooses subgraphs based on the target edge. (B) is the spatial-temporal node encoding module which is responsible for encoding nodes and capturing the global, local, and temporal information hidden in the subgraphs using diffusion-based spatial encoding, distance-based spatial encoding, and relative temporal encoding respectively. The final node encoding is built by combining all previously mentioned encoding. (C) shows the dynamic graph transformer module. This module constructs the edge embedding using a modified multi-headed attention network. Finally, in (D) you can see the discriminative anomaly detector which is trained using samples and pseudo labels generated by a negative sampling strategy. Given that TADDY is only capable of handling non-attributed graphs, we assume that a simplified version of data is utilized to accommodate this limitation.}
                \label{fig:transformers}
\vspace{-10pt}
\end{figure*}
\vspace{-5pt}
\subsubsection{Deep Embedding}
Most of the methods that we have discussed until now are not capable of modeling more complicated dependencies such as complex inter-sensor relationships, but "Deep Encoders" have more capacity and are able to build richer latent representations. Graph Deviation Network (GDN) \cite{deng2021graph} introduced a structure learning approach in which the graph edges are initially unknown and have to be learned. GDN \cite{deng2021graph} is consist of four important components. The first component is Sensor Embedding which captures the characteristics of each node or sensor. In the next step, Graph Structure Learning Learns the complicated relationships between pairs of sensors and models them to the edges of the graph. After the construction of the graph, Graph Attention-Based Forecasting predicts the behavior of sensors in the future time step. Finally, Graph Deviation Scoring will compare these predictions and the actual values in each time step and identify anomaly points that deviate from expected values. AddGraph \cite{ijcai2019-614} is able to aggregate more information such as structural, temporal, and content features to be able to build a more powerful embedding for anomaly detection in dynamic graphs. This model makes use of a graph convolutional neural network (GCN) for capturing structural and content features. By adding Gated Recurrent Units (GRU) with attention modules, AddGraph \cite{ijcai2019-614} makes it possible to combine the long-term and short-term states of each node. Finally, based on the representation that contains structural, temporal, and content features, AddGraph \cite{ijcai2019-614} computes the anomalous score for edges using a single-layer network.

Smart Video Surveillance Systems (Section \ref{sec:smartsurv}), Sensor Network (Section \ref{sec:sensor}), and LAN Network (Section \ref{sec:lan}) all contain many nodes. As a result, Shallow embedding techniques are not adequate for modeling these kinds of networks. We should also consider that all of these three conceptual use-cases are also dynamic (as an example adding a new camera to the Smart Video Surveillance System changes the graph of the network); thus, methods like DeepWalk \cite{10.1145/2623330.2623732}, and Node2Vec \cite{10.1145/2939672.2939754} are not suitable since they are unable to generalize to unseen nodes. Another important issue is that nodes and edges can have features with different schemes that are useful for anomaly detection. Consider the Sensor Network; in each time step, a huge amount of data is generated by a large number of interconnected sensors. The data from each sensor can be related to other sensors with a complex non-linear relationship. Shallow embedding techniques fail to take advantage of this kind of information. Thus, Deep Embedding methods that are more complicated such as GDN \cite{deng2021graph}, and AddGraph \cite{ijcai2019-614} can be helpful in this case. Based on what we discussed, let's use one of the introduced examples discussed in Section \ref{sec:usecases} for better understanding. In the LAN Cluster example, AddGraph seems to be a suitable choice. 

Fig \ref{fig:addgraph} shows the structure of AddGraph which is applied to the LAN Cluster example. Previous hidden states and current snapshot of the network graph is used for constructing the node embedding. Also, an attention module and GRU combine the long-term and short-term states to generate the current hidden state. In the final step, a scoring function is responsible to assign a normality score to the current hidden state. The main points of discussed techniques are available in Table \ref{algorithms}.

\begin{figure*}[h]
        \centering
               \includegraphics[width=0.9\linewidth, height=0.3\linewidth]{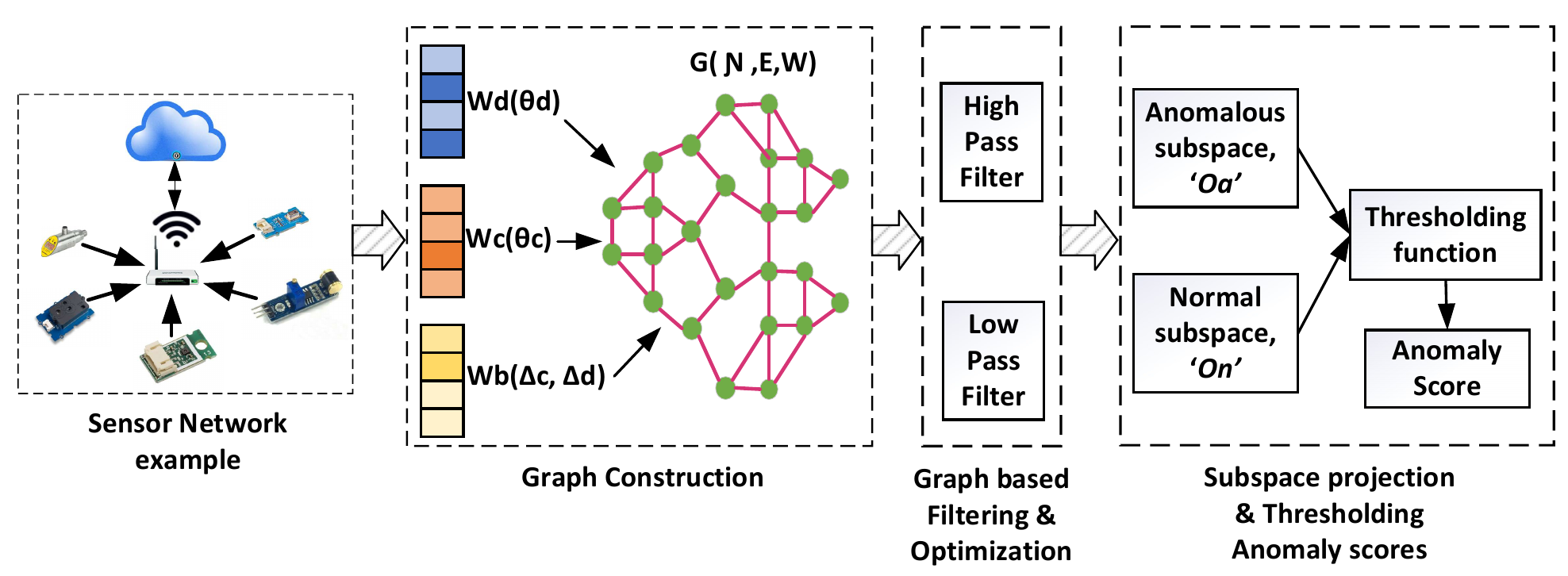}
                \vspace{-10pt}
                \caption{A small part of the Sensor Network introduced in Section \ref{sec:sensor} is used to show Graph Signal Processing for anomaly detection. A 3-stage process that begins with Graph construction from 3 adjacency matrices to generate an undirected, weighted, and connected graph G(N, E, W). Filtering and Optimization methods are next used to get the cut-off frequency of the GSP filter that is further projected as a normal and an anomalous subspace. Both these values in addition to a thresholding function are used to generate the anomaly scores.}
                \vspace{-10pt}
                \label{fig:graphsignalprocessing}
\end{figure*}
\vspace{-10pt}
\subsection{Graph Transformers}
The concept of attention mechanism introduced by \cite{vaswani2017attention} was first used for natural language processing. This mechanism aims to do one-step prediction instead of recurrent processing of the data. By this means, the attention mechanism reduces the path length of the computation, which means reducing the information loss and focusing on the most important features for predicting the output. Adapting this technique to graphs can improve the outcome in many applications, such as anomaly detection. Graph transformer networks are able to automatically generate meta-paths by learning and solving the aforementioned problem enabling many applications such as anomaly detection to work on complex heterogeneous systems \cite{yun2019graph}. In \cite{feng2021heterogeneity}, the model takes advantage of relational graph transformers for finding anomalous nodes in a supervised manner. In this work, first, the heterogeneous graph of the network is extracted, and relational graph transformers and a semantic attention network are used for modeling the complex relationships between the nodes and encoding them, and then classifying them for finding anomalous nodes.

Supervised learning is not possible in many applications due to the lack of labeled data. Using an unsupervised setting, the transformer-based Anomaly Detection framework for Dynamic graphs (TADDY) \cite{9599560} detects anomalous edges in dynamic graphs. TADDY \cite{9599560} consists of four main modules as shown in \figref{fig:transformers}. Edge-based substructure sampling captures the spatial-temporal context of each target edge using the graph diffusion method \cite{klicpera2019diffusion, hassani2020contrastive}. For each edge, this module constructs a fixed-length importance-aware set of neighboring nodes. Then, TADDY \cite{9599560} leverages a novel spatial-temporal node encoding for generating node embedding. After acquiring the embedding from the previous step, a transformer network is used as an encoder to capture spatial and temporal features, followed by a pooling module for aggregating the embedding of all nodes in the same neighboring set. In the final step, the discriminative anomaly detector (which consists of a fully connected layer). 

Graph Learning with Transformer for Anomaly detection (GTA) \cite{chen2021learning}, leverages a modified transformer network. This model detects anomalies on multivariate time series generated from different sensors. These sensors can be related to each other in complicated unknown connections. GTA \cite{chen2021learning} first learns the dependencies using Gumbel-Softmax Sampling \cite{jang2016categorical} strategy. Once this topological structure is established, a graph convolution block updates each node representation by aggregating neighbors' information and message passing to enrich the representations. Now, dilated convolution \cite{yu2015multi} is used for extracting temporal context, but with one novel modification. They have used a hierarchical scheme to make GTA \cite{chen2021learning} capable of capturing temporal patterns with different lengths. On top of these, a more sophisticated and efficient version of transformers is introduced. Multi-branch attention module that is used in GTA \cite{yu2015multi} extracts long-distance temporal dependencies and neighboring nodes' information. In the anomaly scoring module, the original input time series are divided into training sequences (for the encoder) and label sequences (for the decoder). The decoder predicts the behavior of the time series in the target section, and by comparing the predicted output and the actual values, outliers can be detected

Capabilities of all mentioned models can be seen in Table \ref{algorithms}. Unfortunately, none of the anomaly detection models using graph transformers are compatible with attributed graphs which makes them less practical in real applications. Future research can move in the direction of solving this issue to overcome the complex nature of heterogeneous distributed systems. 
\vspace{-10pt}
\subsection{Graph Signal Processing}
Graph Signal Processing (GSP) is another technique that utilizes classical signal processing tools like Fourier transform, filtering, frequency
response etc., to process data defined on both regular and irregular graph networks. \cite{overviewgsp} presents the recent advances in the currently developing GSP tools. Graph signals can be filtered and sampled to apply low-level processing techniques such as denoising and compression. One of the major differences between GSP and traditional Machine learning algorithms is that ML typically considers a graph as a discrete version of a complex network. However, this assumption falls flat for many real-world applications associated with graphs. GSP, on the other hand, looks at existing problems from different perspectives. As an example, defining a graph for sensor-based networks involves choosing edge weights as a decreasing function of the distance between nodes represented by sensors. Observations from similar nodes can lead to a smooth graph function that can detect outliers or abnormal values through high pass filtering or thresholding \cite{dspongsp}. Moreover, a sparse set of sensor readings can also be used to build signal reconstruction methods that can be used to save on resources in sensor networks \cite{jointroutinggsp, spectralcompressgsp}.

\begin{figure*}[!h]
        \centering
               \includegraphics[width=0.8\linewidth, trim= 16 20 16 15,clip]{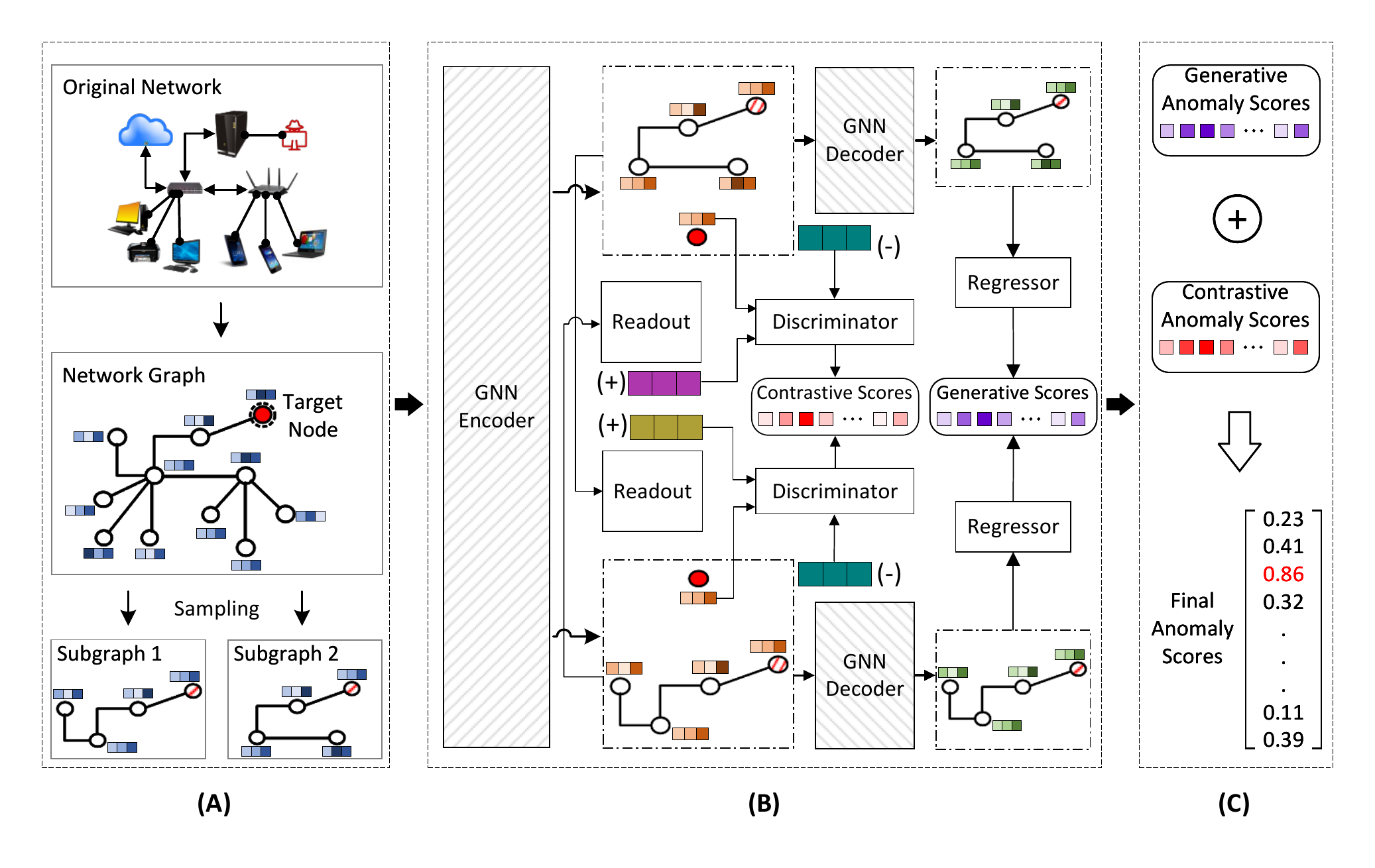}
               \vspace{-10pt}
                \caption{In this figure we have applied the SL-GAD \cite{9568697} frame work to the LAN conceptual use-case discussed in Section \ref{sec:lan}. (A) shows the Graph view sampling module which chooses the target node and samples two subgraphs. (B) is the generative and contrastive discrimination modules. First, the target node and the two subgraphs are fed to the GNN encoder and graph embedding is created. In the next step, two different objectives, the discriminator and regressor try to capture anomalies in the graph structure and attributes. The generative regression module is designed for capturing anomalies in the attributes of each node, while the discriminator module is responsible for finding anomalies in the structure of the graph. Finally, in (C) the contrastive scores and the generative scores are combined to calculate the final anomaly score.}
                \vspace{-10pt}
                \label{fig:contrastive}
\end{figure*}

Anomaly detection with GSP filtering on Wireless Sensor Networks (WSNs) has been a topic of interest in several works. \cite{spectralgsp} captures proximity information such as data between sensors to capture local anomalous behavior. They present three graph designs and use GSP filtering to find the cut-off frequency and Lambda for the filters. This is used to separate normal and anomalous sub-spaces for unsupervised detection. The anomalous space projections are finally utilized to generate anomaly scores as shown in \figref{fig:graphsignalprocessing}. The main advantage of such a system is that in addition to raw sensor data, relational characteristics between the nodes, like proximity information between sensors and their environment, can be effectively captured. Graph-based filtering is found to be particularly useful for both regular and irregular graph structures for unsupervised anomaly detection. In Table \ref{algorithms}, specification of the reviewed model is available. The ability to work on dynamic attributed graphs makes \cite{spectralgsp} aligned with real-world requirements. On the other hand, leveraging GSP techniques can add deeper perspectives to anomaly detection problems and equip them with more relational information helpful for detecting outliers which is essential in real-world applications.  
\vspace{-10pt}
\subsection{Graph Contrastive Learning}
In graph contrastive learning techniques, the objective is to construct a representation by contrasting pairs of data points. The loss function is designed in such a way that by optimizing it, the positive pairs (matching pairs) of data points are brought together, and the negative pairs will be separated. As a result, using this approach, the model is able to learn higher-level representations that are more powerful and distinguishable. In the context of anomaly detection, contrastive learning can be helpful since it is designed to provide a measure of similarity between data pairs. The main idea in GCCAD \cite{chen2021gccad} is to detect anomalies using their distance from average normal points or global context. The key concept is that the outliers will have different features than most points; thus, by contrasting each node with the global context, we can define a measure for detecting anomalies. 

GCCAD \cite{chen2021gccad} has shown that the node embedding made by traditional GNN encoders is not able to discern anomalous nodes properly. However, GCCAD \cite{chen2021gccad} is able to construct a more powerful embedding that highlights the differences between anomalous nodes and the global context. This work uses a context-aware loss function in a supervised manner. The loss function is optimized in the GNN encoder which consists of three modules; edge update, node update, and graph update module. The edge update module is responsible for calculating the likelihood of being a suspicious edge (an edge that connects a normal node to an anomalous one) and removing it, and updating the adjacency matrix. These modules are designed to conserve the homophily assumption between neighboring nodes. This assumption suggests that the neighboring nodes have the same labels. Still, homophily is violated in the case of a connection between a normal node and an anomalous one, and this violation has not been considered in traditional GNN networks. In the next step, the node embedding is updated using message passings by the node update module. Lastly, the graph update module updates the global context. The updated global context is the weighted aggregation of all the nodes. The authors also introduced a self-supervised version of this network named GCCAD-pre \cite{chen2021gccad}. 

\begin{table*}[htbp]
\renewcommand{\arraystretch}{1.15}
  \centering
  \caption{Time complexity of reviewed algorithms. For symbol descriptions please refer to table \ref{tab:symbols}.}
  
    \begin{tabular}{c||cc|c|>{\centering\arraybackslash}m{18em}}
    \textbf{Approach} & \multicolumn{2}{c|}{\textbf{Model}} & \textbf{Time Complexity} & \textbf{Description} \bigstrut[b]\\
    \hline
    \hline
     \multirow{2}[5]{*}{\textbf{Federated Leraning}} & \multicolumn{2}{c|}{DIoT\cite{2019diot}} & - & - \bigstrut[b]\\
\cline{2-5}       & \multicolumn{2}{c|}{VFL \cite{vfl2021}} &  $\mathcal{O}\left(n {m}^{2}\right)$ & - \bigstrut\\
\hline
    \multirow{9}[17]{*}{\textbf{Auto Encoders}} & \multicolumn{2}{c|}{\cite{borghesi2019anomaly}} & -  & - \bigstrut[b]\\
\cline{2-5}       & \multicolumn{2}{c|}{\cite{8363930}} & -  & - \bigstrut\\
\cline{2-5}       & \multicolumn{2}{c|}{\cite{8532358}} & -  & - \bigstrut\\
\cline{2-5}       & \multicolumn{2}{c|}{RDA  \cite{10.1145/3097983.3098052}} & -  & - \bigstrut\\
\cline{2-5}       & \multicolumn{2}{c|}{ITSR \cite{10.1007/978-3-030-46150-8_13}} & -  & - \bigstrut\\
\cline{2-5}       & \multicolumn{2}{c|}{\cite{CGN_particle}} & -  & - \bigstrut\\
\cline{2-5}       & \multicolumn{2}{c|}{DONE \cite{bandyopadhyay2020outlier}} & $\mathcal{O}(Nd)$ & - \bigstrut\\
\cline{2-5}       & \multicolumn{2}{c|}{AdOne \cite{bandyopadhyay2020outlier}} & $\mathcal{O}(Nd)$ & - \bigstrut\\
\cline{2-5}       & \multicolumn{2}{c|}{DeepSphere \cite{teng2018deep}} & -  & - \bigstrut\\
    \hline
    \multirow{9}[17]{*}{\textbf{Graph Embedding}} & \multicolumn{1}{c|}{\multirow{7}[14]{*}{Shallow Encoders}} & Deep Walk \cite{10.1145/2623330.2623732} & -  & - \bigstrut\\
\cline{3-5}       & \multicolumn{1}{c|}{} & Node2Vec \cite{10.1145/2939672.2939754} & $\mathcal{O}(\frac{l}{s(l-s)})$ & Per sample complexity \bigstrut\\
\cline{3-5}       & \multicolumn{1}{c|}{} & LINE \cite{10.1145/2736277.2741093} & $\mathcal{O}(m\hat{n}\hat{d})$ & - \bigstrut\\
\cline{3-5}       & \multicolumn{1}{c|}{} & TADW \cite{yang2015network} & - & - \bigstrut\\
\cline{3-5}       & \multicolumn{1}{c|}{} & \multirow{3}[6]{*}{NetWalk \cite{10.1145/3219819.3220024}} & $\mathcal{O}(nl|\Omega|)$ & The complexity of walk generation  \bigstrut\\
\cline{4-5}       & \multicolumn{1}{c|}{} &    & $\mathcal{O}(md)$ & The complexity of edge encoding \bigstrut\\
\cline{4-5}       & \multicolumn{1}{c|}{} &    & $\mathcal{O}(cd)$ & The complexity of anomaly detection on incoming data points \bigstrut\\
\cline{2-5}       & \multicolumn{1}{c|}{\multirow{2}[3]{*}{Deep Encoders}} & GDN \cite{deng2021graph} & -  & - \bigstrut\\
\cline{3-5}       & \multicolumn{1}{c|}{} & AddGraph \cite{ijcai2019-614} & -  & - \bigstrut[t]\\
\hline
    \multirow{3}[5]{*}{\textbf{Graph Transformers}} & \multicolumn{2}{c|}{\cite{feng2021heterogeneity}} & -  & - \bigstrut[b]\\
\cline{2-5}       & \multicolumn{2}{c|}{TADDY \cite{9599560}} &  $\mathcal{O}\left(\tau k m I+T \tilde{n}^{2}\right)$ & - \bigstrut\\
\cline{2-5}       & \multicolumn{2}{c|}{GTA  \cite{yu2015multi}} & $\mathcal{O}(4\tau D ^{2}+2\tau^{2} D )$ & This is for the simplest attention module. More complex ones are available in the original paper.  \bigstrut\\
    \hline
        
 {\textbf{Graph Signal Processing}} & \multicolumn{2}{c|}{\cite{spectralgsp}}     &   -   & \bigstrut\\

    \hline
    \multirow{3}[4]{*}{\textbf{Graph Contrastive Learning}} & \multicolumn{2}{c|}{GCCAD \cite{chen2021gccad}} & $\mathcal{O}(m)$ & - \bigstrut[b]\\
\cline{2-5}       & \multicolumn{2}{c|}{CoLA \cite{liu2021anomaly}} & $\mathcal{O}(kn\hat{R}(\eta+c))$ & - \bigstrut\\
\cline{2-5}       & \multicolumn{2}{c|}{SL-GAD \cite{9568697}} & $\mathcal{O}(Rnk(\eta+K))$ & - \bigstrut[t]\\
    \end{tabular}%
  \label{tab:complexity}%
  \vspace{-10pt}
\end{table*}%

CoLA \cite{liu2021anomaly} focuses on anomaly detection in large-scale attributed graphs, which is applicable to real-world problems. This novel network has three main components: pair sampling, a GNN-based contrastive learning model, and the score computation module. The instance pair sampling is designed to generate the pairs of data for the training phase. In contrast with GCCAD \cite{chen2021gccad} that contrasts nodes with the global context, the definition of pairs in CoLA \cite{liu2021anomaly} focuses on the relationship between a node and its neighbors. Hence the strategy for defining the pairs is "target node versus local subgraph". The GNN-based contrastive learning module consists of three sub-modules: (1) GNN model, (2) readout module, and (3) discriminator module. After acquiring the pairs, the GNN module will extract embedding for the target node and the local subgraph. Then the readout module uses the average pooling function to create a vector of embedding of all nodes in the embedded subgraph. In the discriminator module, the positive and negative pairs are contrasted, and a score is generated. Finally, the score computation module measures the anomaly score for all nodes, and anomalous nodes can be picked by choosing the nodes with the highest scores. SL-GAD \cite{9568697}, introduced a novel generative and contrastive self-supervised model. 

SL-GAD \cite{9568697} is made of three major components similar to CoLA \cite{liu2021anomaly}; graph view sampling, contrastive self-supervised learning, and graph anomaly scoring. In the graph view sampling, for each target node, two local subgraphs are extracted. Next, in the contrastive self-supervised learning module, a GNN constructs the latent representation of the samples. In addition to contrastive scores, SL-GAD \cite{9568697} also leverages a graph autoencoder network for reconstructing the feature vector of the target node in order to fully utilize the contextual information of target nodes. Finally, graph anomaly scoring predicts the final scores based on both generative scores (from the graph autoencoder) and contrastive scores (from contrastive learning).  

\figref{fig:contrastive} shows the work flow of the SL-GAD \cite{9568697} applied to the LAN Cluster example introduced in Section \ref{sec:lan}. The sampled subgraphs are fed to the GNN Encoder and the encoded representations are fed to two different branches, the discriminative and the generative modules. at the final step, anomaly scores from these two branches are combined to construct the final anomaly score.

Characteristics of all mentioned models can be seen in Table \ref{algorithms}. As discussed, Graph Contrastive Learning approaches show promising results and are extremely adaptable to real-world scenarios. The discussed works are all able to overcome the dynamic complex nature of distributed heterogeneous systems and the unsupervised/self-supervised learning adaptability of these models can alleviate the problem of available labeled data mentioned in Section \ref{sec:riskreq}.  




\begin{table}[htbp]
  \centering
  \caption{Table \ref{tab:complexity} symbol description.}
    \begin{tabular}{P{0.2\columnwidth}|P{0.7\columnwidth}}
    Parameter & Description \bigstrut[b]\\
    \hline
    \hline
    n  & number of nodes in the graph \bigstrut\\
    \hline
    $\tilde{n}$ & average number of nodes in a timestamp \bigstrut\\
    \hline
    m  & number of edges in the graph \bigstrut\\
    \hline
    k  & number of nodes in a subgraphs \bigstrut\\
    \hline
    T  & number of timestamps \bigstrut\\
    \hline
    $\tau$ & size of the time window \bigstrut\\
    \hline
    $\hat{n}$ & number of negarive samples \bigstrut\\
    \hline
    l  & length of a walk \bigstrut\\
    \hline
    $|\Omega|$ & number of walks \bigstrut\\
    \hline
    s  & number of samples \bigstrut\\
    \hline
    D & number of input dimension \bigstrut\\
    \hline
    d  & latent dimension of node representation \bigstrut\\
    \hline
    $\hat{d}$ & degree of a node \bigstrut\\
    \hline
    c  & number of clusters \bigstrut\\
    \hline
    R  & number of evaluation rounds \bigstrut\\
    \hline
    I  & number of iterations \bigstrut\\
    \hline
    $\hat{R}$ & number of sampling rounds for each node \bigstrut[t]\\
    \end{tabular}%
  \label{tab:symbols}%
  \vspace{-10pt}
\end{table}%

\vspace{-10pt}
\section{Comparison and Discussion}
\label{sec:discuss}

In this section, first, we discuss the metrics that are available for evaluation and then we try to compare the methods discussed in Section \ref{sec:graphs} from an algorithmic perspective. Then we discuss the requirements and challenges that utilizing graphs adds to previously mentioned requirements discussed in Section \ref{sec:riskreq}, and finally, we discuss the future directions that research in this field can take.
\vspace{-10pt}
\subsection{Metrics and Comparison}
\label{sec:metrics}
The performance and viability of any anomaly detection algorithm, graph-based or non-graph-based, can be evaluated by Precision, Recall, Accuracy, Receiver Operating Characteristic (ROC) curve, and Area Under the ROC Curve (AUC). While Precision, Recall, and Accuracy are a measure of the true positives, false positives, true negatives, and false negatives, AUC, on the other hand, summarizes the information contained in the ROC curve. Larger AUC values indicate better performance at distinguishing between anomalous and normal observations. 

A major challenge while reporting the above-mentioned metrics is the absence of unified evaluation criteria amongst the diversity of algorithms, the application type, and the heterogeneous nature of data and devices involved. As an example, if we want to compare the latency of the two algorithms, we have to make sure that they are both tested on particular hardware, they are getting tested on a specific job, and we have to eliminate different variables that have an effect on that metric. 

In contrast, \emph{Time Complexity} is The only metric that can give us some insight into the algorithm's performance. Table \ref{tab:complexity} shows the time complexity of reviewed algorithms. In order to assess how well an algorithm performs compared to other algorithms, and also with respect to points in Section \ref{sec:riskreq}, we can refer to Time Complexity. It gives us a good understanding of how fast and efficient the method is, a general idea of the algorithm's complexity, and whether it is suitable for specific tasks or hardware based on the power it needs.

\vspace{-5pt}
\subsection{Graph-specific Challenges}

Graphs are extremely powerful for catching relational information, especially in distributed systems. They help with addressing challenges and meeting the requirements discussed in Section \ref{sec:riskreq}, as well as capturing important features of the data like relational features and information. However, there is a cost to all these benefits. While utilizing graphs these challenges should be addressed:

\textbf{Types of Graphs:} Each system can be represented by a specific type of graph. Each type of graph has its features and is suitable for a group of domains. For example, attributed graphs add more complexity by introducing the features of the connection between the nodes, and the features of the nodes themselves. Also, a graph representing a system might be dynamic or static, depending on the nature of the original system. All these sub-classes of graphs add to the complexity of graph-based learning techniques. Subsequently, the introduced models should be able to handle these diverse types of network graphs, but as we discussed in previous sections, many of the models can just work on plain static graphs which is not satisfactory in many applications. \cite{zheng2018fraudne, dou2021user, liu2017accelerated, zhang2019robust, ding2019deep, wang2021one}

\textbf{Graph Anomalies: } By introducing graphs, we get a better representation of a system. Graphs and Graph Neural Networks make us capable of modeling complicated dependencies in real networks. But if we look at it from another perspective, this adaptation of graph representation introduces different places for anomalies to take place. In graphs, outliers can occur in nodes, edges, subgraphs, and even the whole graph. But models are usually only able to detect one kind of foresaid anomalies, and there is no unique framework to address them all together. This means that the approaches must specifically determine what is the objective of their detection, which types they handle, and where they show the best performance. \cite{ding2019interactive, peng2020deep, li2019specae}

\vspace{-9pt}
\subsection{Future Research Directions}
The most important research direction at the moment seems to be finding a means of quantifying the results of anomaly detection algorithms in a unified manner, thereby enabling comparison in various domains and fields. A great share of all the advancements in technology originates from quantifying and comparing new findings with already existing works. However, considering the variety in this field, comparing algorithms to each other is nearly impossible. First, the best metric for evaluating a model must be identified. Alongside that, an evaluation mechanism, general enough to be applicable to disparate anomaly detection algorithms is also needed. As an example, in computer vision, COCO \cite{10.1007/978-3-319-10602-1_48} provides such an environment for emerging algorithms. A framework of testing can also be another fruitful tool. In this direction, creating a simulator, or a system that can serve as a toy sample of a large distributed system can be beneficial as well. To recapitulate, there is a need for a unified evaluation of graph-based and non-graph-based anomaly detection algorithms and techniques. Following this path, an extension that this survey can benefit from is to include experimental results, by conducting a unified benchmarking. This aspect is particularly promising, as incorporating experimental analysis alongside algorithmic and conceptual analysis can yield a more comprehensive and nuanced understanding of the subject matter.

As discussed in the Challenges Section, the second-largest issue in the way of anomaly detection is the lack of a suitable dataset. Generating or preparing a proper dataset can be a substantial contribution to the field of anomaly detection. However, in the context of anomaly detection, usually, an expert must supervise the process of labeling and collecting data and this fact makes the generated dataset prone to human errors. All of these are added to the actual cost of creating such a dataset. Synthetic datasets are another option, and creating them can be explored through a whole line of research.

Numerous methods of anomaly detection are designed to discover whether an anomaly has happened or not. They are able to notify a responsible person or group to look for further investigation. However, they are not able to identify what sort of anomaly has occurred. A major subject for further work and research is the interpretability of anomalies. Down the road, there is a tendency to distinguish between anomalies using the algorithms at hand. For example, in the Sensor Network mentioned in \ref{sec:sensor} we like the model to be able to distinguish between a cyber attack and power shortage and raise the correct alarm. If we go further than that, the ways that the model can restore the system or stop the anomaly can be investigated.

\vspace{-10pt}
\section{Conclusion}

In this survey, we have discussed several state-of-the-art graph-based approaches for anomaly detection in details. We introduce three unique conceptual use-cases based on real-time complex distributed systems. Our work builds on the existing traditional machine learning and deep learning paradigms used for anomaly detection and goes on to introduce four separate categories of graph neural network algorithms. We further apply our formulated models to each of the methods and provide a thorough review and summarization of the categories. At last, we comprehensively look at graph specific big-data challenges and provide a couple of future research directions in this field.

\vspace{-9pt}
\ifCLASSOPTIONcompsoc
  \section*{Acknowledgments}
\else
  \section*{Acknowledgment}
\fi

The authors would like to thank Siemens Technology for supporting this work.

\ifCLASSOPTIONcaptionsoff
  \newpage
\fi



%
\vspace{-10pt}
\bibliographystyle{IEEEtran}
\bibliography{Mendeley.bib}

%
\vspace{-40pt}
\begin{IEEEbiography}[{\includegraphics[width=1in,height=1.25in,keepaspectratio]{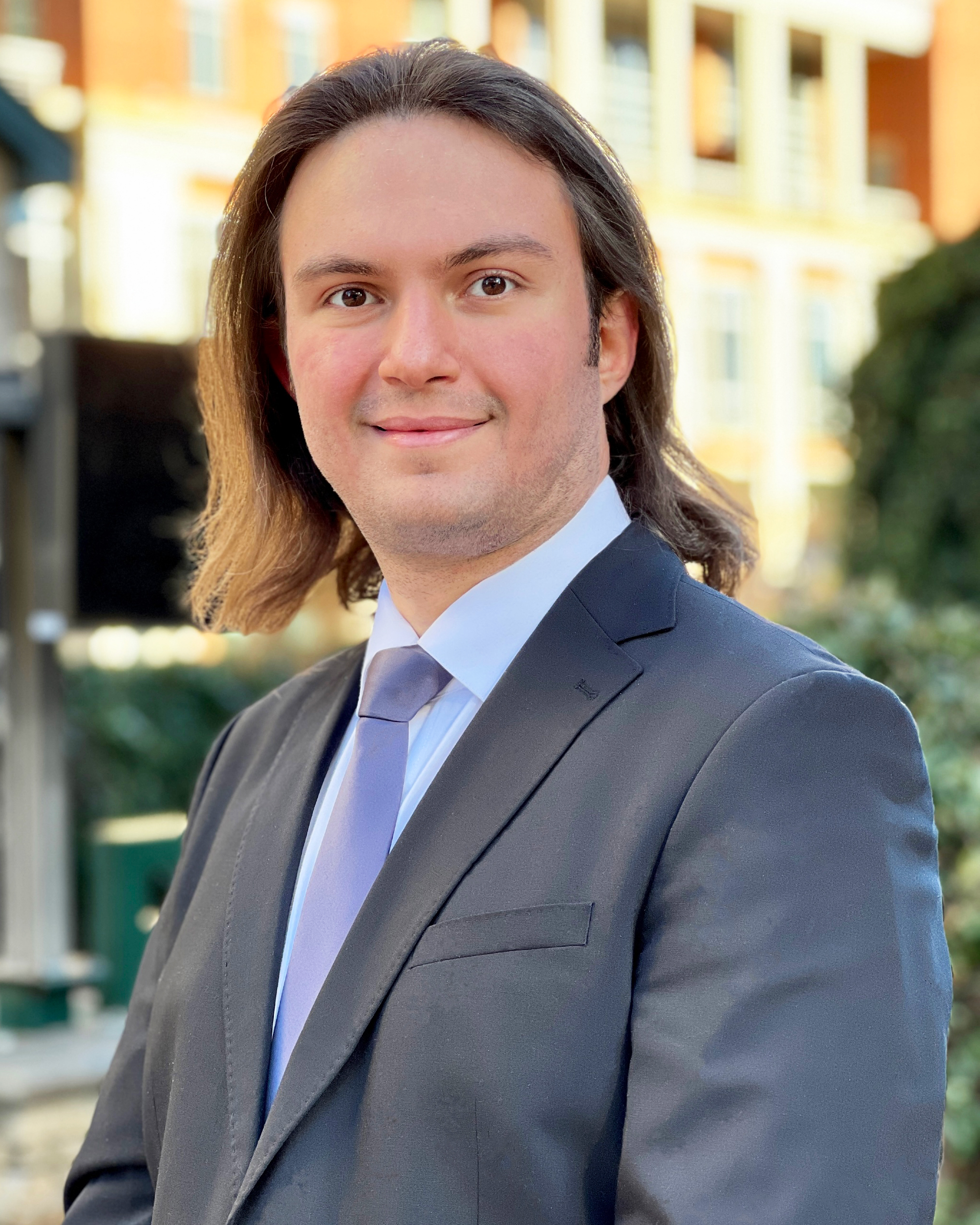}}]{Armin Danesh Pazho} (S’22) is currently a Ph.D. student at the University of North Carolina at Charlotte, NC, United States. With a focus on Artificial Intelligence, Computer Vision, and Deep Learning, his research delves into the realm of developing AI for practical, real-world applications and addressing the challenges and requirements inherent in these fields. Specifically, his research covers action recognition, anomaly detection, person re-identification, human pose estimation, and path prediction.
\end{IEEEbiography}

\begin{IEEEbiography}[{\includegraphics[width=1in,height=1.25in,keepaspectratio]{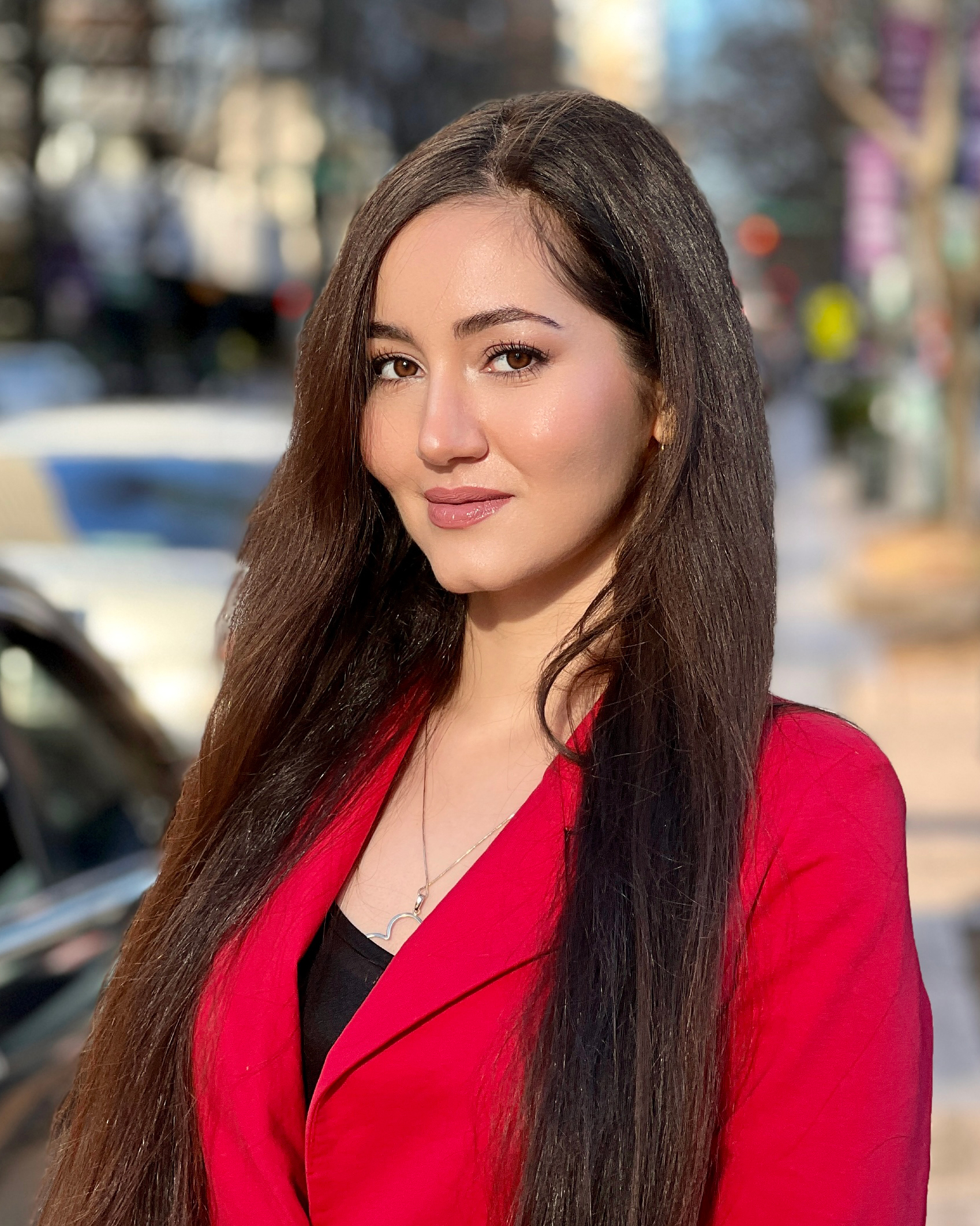}}]{Ghazal Alinezhad Noghre} (S’22) is currently pursuing her Ph.D. in Electrical and Computer Engineering at the University of North Carolina at Charlotte, NC, United States. Her research concentrates on Artificial Intelligence, Machine Learning, and Computer Vision. She is particularly interested in the applications of anomaly detection, action recognition, and path prediction in real-world environments, and the challenges associated with these fields.
\end{IEEEbiography}

\vspace{-30pt}
\begin{IEEEbiography}[{\includegraphics[width=1in,height=1.25in,keepaspectratio]{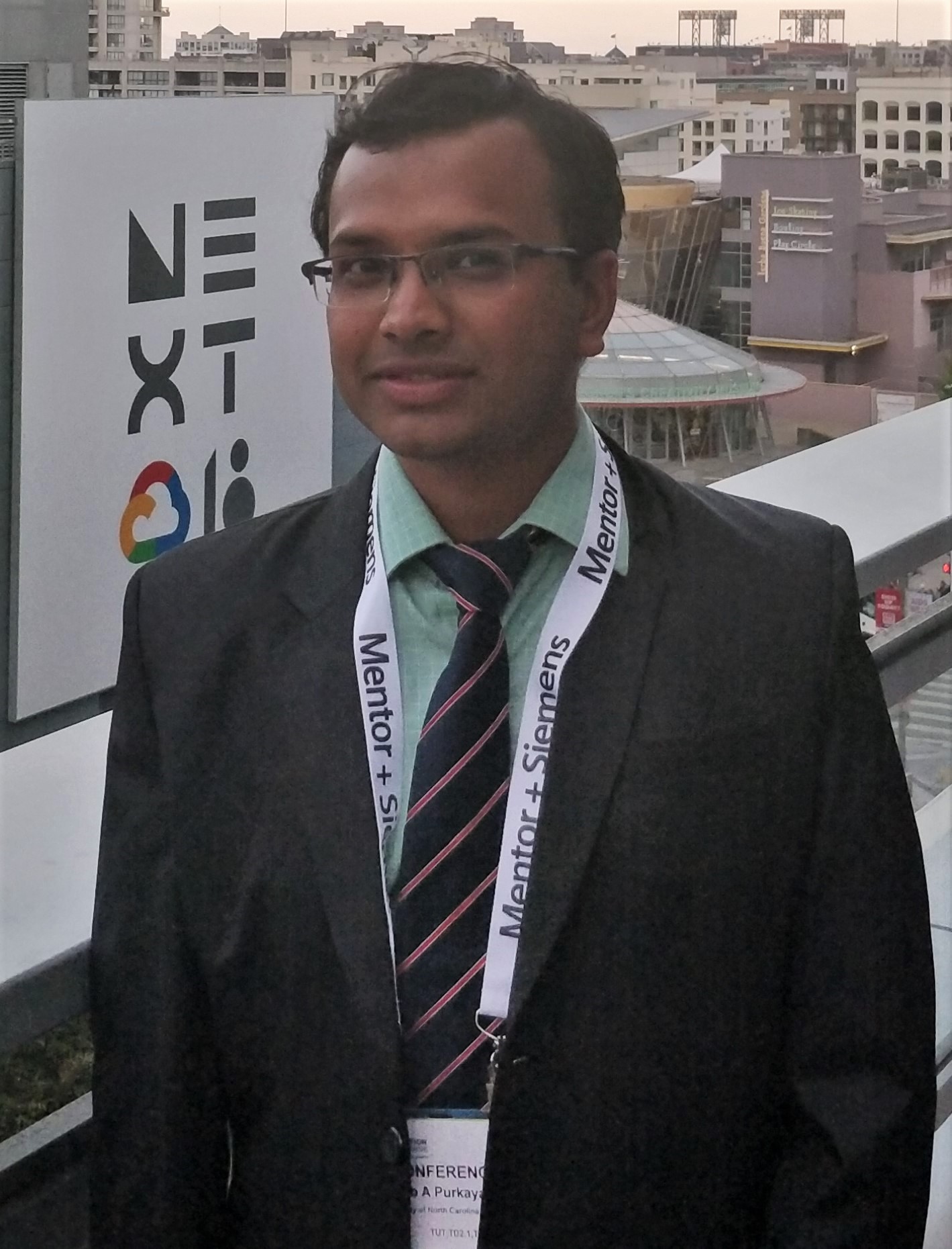}}]{Arnab A Purkayastha} is an Assistant Professor of Electrical and Computer Engineering department at the Western New England University, Massachusetts. He received his PhD in the year 2021 from the University of North Carolina at Charlotte. His research interests and activities lie in the recent advances in High Performance Computing and Machine Learning fields, including system level integration at the cloud and the edge.
\end{IEEEbiography}

\vspace{-40pt}
\begin{IEEEbiography}[{\includegraphics[width=1in,height=1.25in,keepaspectratio]{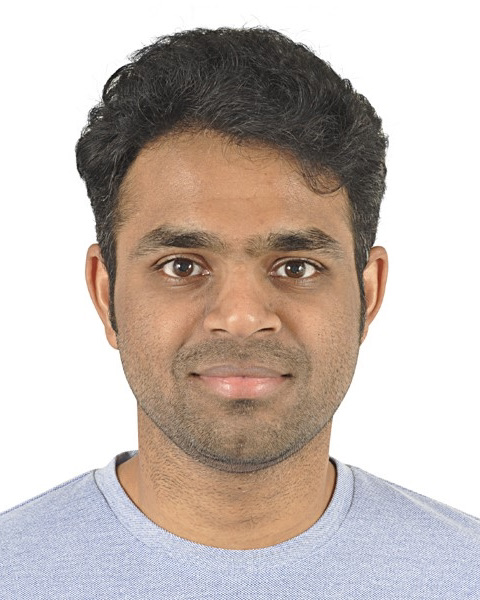}}]{Jagannadh Vempati} is a research scientist at Siemens Technology in Princeton – NJ performing research and development on designing cyber defense solutions that incorporate domain know-how and the semantic context of attack detection and response solutions. His responsibilities include developing new methods and techniques that improve the value of security analytics in Cyber Defense by integrating industrial domain expertise and semantics/context and guiding Siemens Business Units to include research results into products, solutions, and services. His research is focused on data-driven cyber security and designing resilient networks. His research also focuses on designing, developing, and optimizing anomaly detection models using artificial intelligence. He holds a Ph.D. in computer science and engineering from the University of North Texas.
\end{IEEEbiography}

\vspace{-40pt}
\begin{IEEEbiography}[{\includegraphics[width=1in,height=1.25in,keepaspectratio]{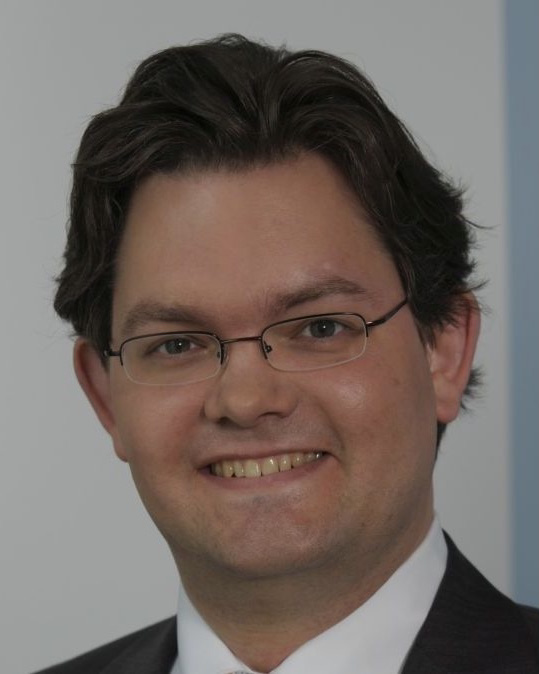}}]{Martin Otto} is a researcher and research manager with Siemens Technology, Siemens' central R\&D organization, since 2005. He is currently the head of the research group "Cybersecurity Service Innovation" at Siemens Corporation, Siemens Technology, in Princeton, NJ, USA. His mission is to provide Siemens business units with technology solutions and innovations that enable Siemens to provide state of the are security services to customers. A specific focus is on helping Siemens customers that operate energy systems and other parts of national critical infrastructure to detect, react to, mitigate, and otherwise defend against cyber attacks. He held positions in the US, Canada, and Germany, among them as global Head of the Siemens CERT (Computer Emergency Response Team). He acquired a Ph.D. in computer Science in 2005 from Paderborn University, Paderborn, Germany, working on fault attack side channels on smart cards. 
\end{IEEEbiography}

\vspace{-40pt}
\begin{IEEEbiography}[{\includegraphics[width=1in,height=1.25in,keepaspectratio]{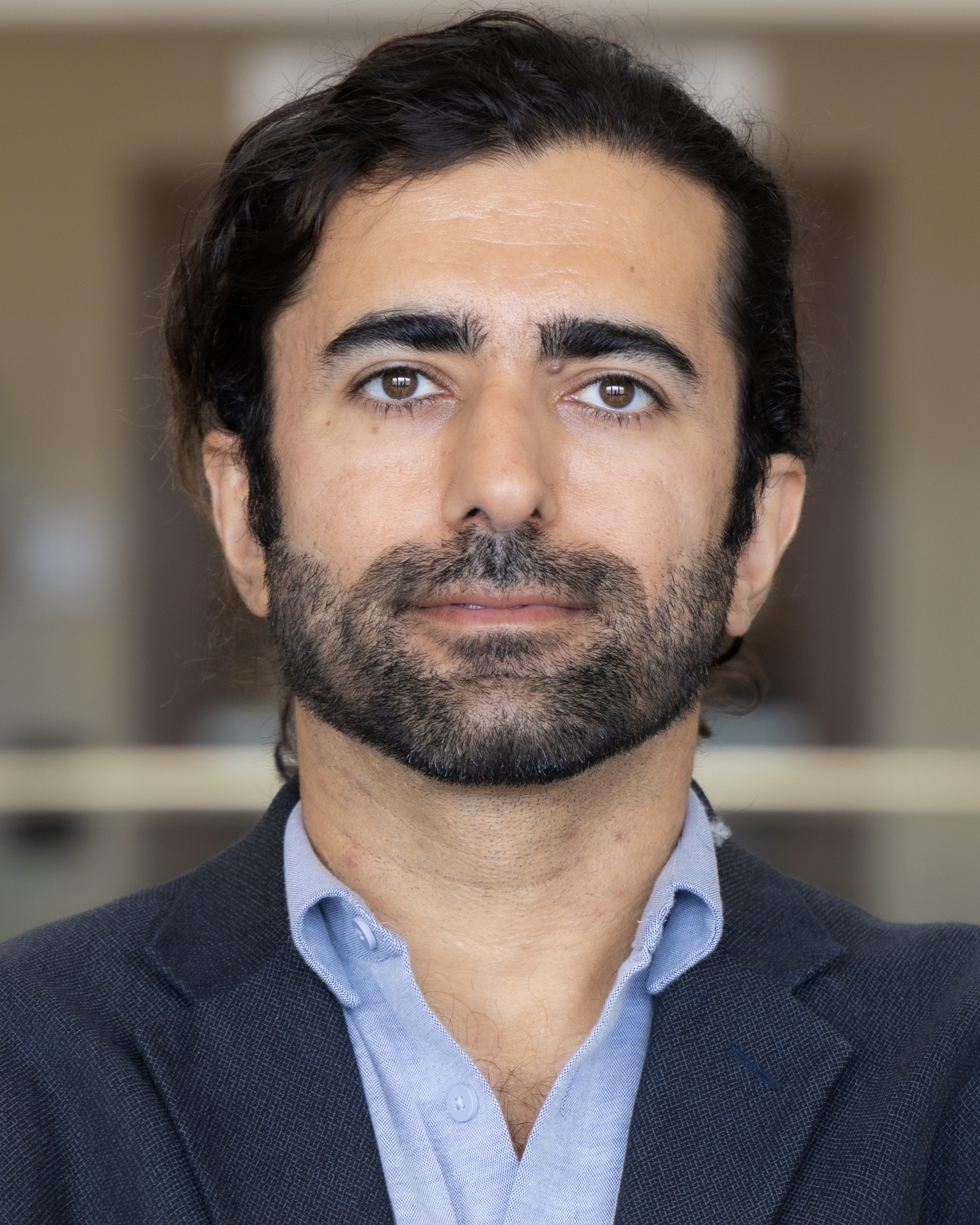}}]{Hamed Tabkhi} (S’07–M’14) is the associate professor of Computer Engineering at the University of North Carolina Charlotte (UNCC). He received his PhD in Computer Engineering from Northeastern University in 2014. His research and scholarship activities focus on transformative computer system solutions to bring recent advances in Artificial Intelligence (AI) to address real-world problems. In particular, he focuses on AI-based solutions to enhance our communities' safety, health, and overall well-being.
\end{IEEEbiography}







\end{document}